\documentclass[10pt,twocolumn,letterpaper]{article}

\usepackage[final]{cvpr}      
\usepackage{algorithm}
\usepackage{algorithmicx}
\usepackage{algpseudocode}
\usepackage{multirow}
\usepackage{array}
\usepackage{amsmath} 
\usepackage{amssymb} 
\usepackage{booktabs}
\usepackage{siunitx} 
\usepackage{graphicx} 
\definecolor{cvprblue}{rgb}{0.21,0.49,0.74}
\usepackage[pagebackref,breaklinks,colorlinks,allcolors=cvprblue]{hyperref}

\def\paperID{} 
\def\confName{CVPR}
\def\confYear{2026}

\newcommand{\wjx}[1]{{\textcolor{black}{#1}}}

\newcommand{\zmz}[1]{{\textcolor{black}{#1}}}
\newcommand{\jwx}[1]{{\textcolor{black}{#1}}}

\title{StegoNGP: 3D Cryptographic Steganography using Instant-NGP}

\author{Wenxiang Jiang\\
Shandong Technology and Business University\\
{\tt\small jwx@sdtbu.edu.cn,}
\and
Yujun Lan\\
Ocean University of China\\
{\tt\small lanyujun4733@stu.ouc.edu.cn}
\and
Shuo Zhao\\
Ocean University of China\\
{\tt\small zhaoshuo@ouc.edu.cn}
\and
Yuanshan Liu\\
Shandong Technology and Business University\\
{\tt\small yshliu@sdtbu.edu.cn}
\and
Mingzhu Zhou $^*$\\
Adelaide University\\
{\tt\small mingzhu.zhou@student.adelaide.edu.au}
\and
Jinxin Wang $^*$\\
Shandong Technology and Business University\\
{\tt\small wangjinxin@sdtbu.edu.cn}
}

\begin{document}
\maketitle

\begin{abstract}
\wjx{Recently, Instant Neural Graphics Primitives (Instant-NGP)}
has achieved significant success in rapid 3D scene reconstruction, but securely embedding high-capacity hidden data, such as an entire 3D scene, remains a challenge. Existing methods rely on external decoders, require architectural modifications, and suffer from limited capacity, which makes them easily detectable.
\wjx{We propose a novel parameter-free \emph{3D Cryptographic \textbf{Steg}an\textbf{o}graphy using Instant-\textbf{NGP} (StegoNGP)}, which leverages the Instant-NGP hash encoding function as a key-controlled scene switcher.} 
By associating a default key with a cover scene and a secret key with a hidden scene, our method trains a single model to interweave both representations within the same network weights. The resulting model is indistinguishable from a standard Instant-NGP in architecture and parameter count. We also introduce an enhanced Multi-Key scheme, which assigns multiple independent keys across hash levels, dramatically expanding the key space and providing high robustness against partial key disclosure attacks.
Experimental 
\wjx{results demonstrated that} StegoNGP can hide a complete high-quality 3D scene with 
\jwx{strong}
imperceptibility and security, \wjx{providing} a new paradigm for high-capacity, undetectable information hiding in neural fields. The code can be found at [link will be publicly available upon acceptance].
\end{abstract}    
\section{Introduction}

\begin{figure}[bt]
  \centering
  \includegraphics[width=0.47\textwidth] {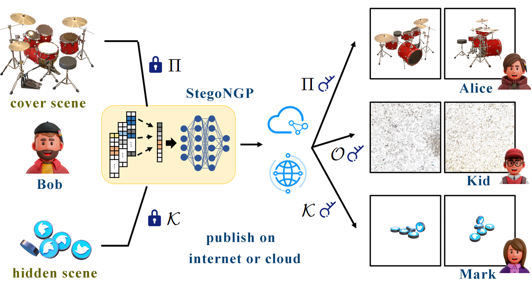}
  \caption{Application scenario of the proposed StegoNGP: A 3D model creator (Bob) creates a 3D scene (cover scene) and embeds hidden information (e.g., copyright information) into it using a steganographic key $\mathcal{K}$. The trained StegoNGP is then publicly released on the internet or a cloud drive, and its model is indistinguishable from a standard Instant-NGP model. A normal user (Alice) can \wjx{utilize} and render the original cover scene. Mark, the intended recipient holding the key $\mathcal{K}$, can extract the hidden scene, thereby achieving copyright protection or secret communication. Kid, who attempts to access the hidden scene with an irrelevant key $\mathcal{O}$, will only obtain meaningless blank or noise images.}
  \label{fig:intro}
\end{figure}

Neural Radiance Fields (NeRF) \cite{mildenhall2021nerf} have significantly advanced the field of 3D reconstruction by implicitly learning 3D representation using Multilayer Perceptrons (MLP). Instant Neural Graphics Primitives (Instant-NGP) \cite{muller2022instant} and 3D Gaussian Splatting (3DGS) \cite{kerbl20233d} have further enhanced real-time rendering, fidelity, and usability by employing hybrid explicit-implicit representations. 
As the \zmz{repository} of these 3D models grows, 
their value as digital assets has rapidly escalated. Consequently, there is a growing demand for securely and imperceptibly hiding high-capacity information for purposes such as cryptographic copyright protection, traceable metadata embedding, and secret communication. This has spurred research in fields like 3D steganography \cite{li2025gs, jang2024waterf, sun2024rwnerf} and watermarking \cite{li2023steganerf, dong2023steganography, zhang2024gs}.

Traditionally, most NeRF watermarking techniques have focused on embedding low-capacity data, such as binary messages \cite{song2024protecting, chen2025guardsplat}, into the model for copyright protection. In contrast, NeRF steganography methods, such as StegaNeRF \cite{li2023steganerf} and NeRF Steganography \cite{dong2023steganography} 
\wjx{can} embed a 2D image into the 3D model. However, these methods suffer from limited capacity, require external decoders, and are 
\jwx{unable to support secure communication}. IPA-NeRF \cite{jiang2024ipa} and GaussTrap \cite{hong2025gausstrap} \wjx{utilize} a secret viewpoint as a backdoor trigger to embed an image at a specific view in NeRF and 3DGS models, but this approach also has limited capacity and effectively causes damage to the 3D model's integrity.

Hiding high-capacity 
\wjx{encrypted} information within compact 3D neural radiance field representations, such as Instant-NGP \cite{muller2022instant}, while maintaining communication privacy and security, remains a significant challenge, especially without altering the 3D model architecture or adding extra parameters or decoders.

In this paper, we propose 
\wjx{a \emph{3D cryptographic \textbf{steg}an\textbf{o}graphy using Instant-\textbf{NGP} (StegoNGP)}} to address this challenge. We transform the Instant-NGP hash mapping function into a key-controlled scene switcher. The original hash function in Instant-NGP relies on a default sequence of large prime numbers to map 3D coordinates to the feature table. We treat this prime sequence as the default key $\Pi$. We then introduce a new secret prime sequence $\mathcal{K}$, as the steganographic key. A single Instant-NGP model is trained to simultaneously optimize two scenes: the cover scene $\mathcal{S}$ using key $\Pi$, and the hidden scene $\mathcal{B}$ using key $\mathcal{K}$. This enables us to hide a complete 3D scene within another scene's model with zero parameter overhead, accessible only via the cryptographic key. The application scenario is shown in Figure \ref{fig:intro}.

Notably, 
GS-Hider \cite{zhang2024gs} achieves 3D scene steganography based on the 3DGS model \cite{kerbl20233d} by adding extra spherical harmonics parameters to each anisotropic Gaussian and using an image decoder. In contrast, our research does not add extra parameters or structure to Instant-NGP. This makes the model produced by our method, once publicly released, indistinguishable from a standard Instant-NGP. 
\wjx{In summary}, our work advances the field in the following aspects:

\begin{itemize}
\item We propose StegoNGP, a novel parameter-free 3D cryptographic steganography framework. By exploiting the inherent structural redundancy of hash-based neural fields, our method achieves high-capacity data hiding without requiring auxiliary decoders or architectural modifications.
\item We 
\wjx{devise} a unique key-controlled scene-switching mechanism that uses a secret prime number sequence to hide a complete, high-quality 3D scene. 
\wjx{This} method achieves 
\jwx{strong}
imperceptibility, as the cover scene's quality is unaffected, while security is guaranteed by the key's confidentiality.
\item We introduce a more robust scheme Multi-Key StegoNGP, which assigns multiple independent keys to different hash levels. This dramatically expands the key space and enhances security, proving highly resistant to partial key disclosure attacks, as an attacker must possess the complete key set to recover the hidden scene.
\end{itemize}

\section{Related Work}

\subsection{Neural 3D Scene Representations}

Neural Radiance Fields (NeRF) \cite{mildenhall2021nerf} learn a continuous, implicit 3D representation using MLP networks from a set of 2D images, and are capable of synthesizing photo-realistic new views from unseen viewpoints. 
\jwx{Building on NeRF, subsequent research has rapidly advanced the field, achieving notable progress in multiple directions.} Instant Neural Graphics Primitives (Instant-NGP) \cite{muller2022instant} introduced a multi-resolution 3D hash feature embeddings method, which maps 3D spatial coordinates to trainable feature vectors in multi-level feature hash tables. The concatenated multi-level features are then decoded by a small MLP into color and density, greatly accelerating training and rendering speeds. Meanwhile, 3D Gaussian Splatting (3DGS) \cite{kerbl20233d}, as another improvement for explicitly representing 3D scenes, \wjx{utilizes} millions of 3D Gaussians to represent the scene, also achieving high rendering quality and real-time rendering speeds. Other related work has achieved improvements in reducing model complexity \cite{lindell2021autoint,rebain2021derf}, improving rendering quality \cite{barron2021mip, barron2022mip}, scalability to larger scenes \cite{tancik2022block, Driess2023RLNERF}, and even extending NeRF to dynamic scenes \cite{pumarola2021d, wu20244d}.

\subsection{NeRF Watermarking and Steganography}

As 3D reconstruction technology continuously improves, the models generated by NeRF and its variants are accumulating value as digital assets. Consequently, the demand for copyright protection, metadata embedding, and information hiding has grown. This has spurred research in the two related fields of NeRF watermarking \cite{li2025gs, jang2024waterf, song2024protecting, sun2024rwnerf} and steganography \cite{li2023steganerf, dong2023steganography, zhang2024gs}.

\paragraph{NeRF Watermarking.} NeRF Watermarking primarily focuses on copyright protection and ownership verification. These methods are typically dedicated to embedding low-capacity, robust information, such as binary messages \cite{jang2024waterf, song2024protecting, chen2025guardsplat, jang20253d} or small logos \cite{sun2024rwnerf}. These watermarks are designed to be identifiable even after the model undergoes transformations like compression or fine-tuning. The capacity of such methods is generally small, 
\wjx{such as} 16 bits \cite{jang2024waterf}, 48 bits \cite{song2024protecting, chen2025guardsplat} or 64 bits \cite{jang20253d} of binary information, and requires specialized decoders, which significantly limits the upper bound for transmitting secret information.

\paragraph{NeRF Steganography.} In contrast, NeRF Steganography aims to pursue high capacity and imperceptibility, $i.e.$, hiding large amounts of data without raising suspicion. StegaNeRF \cite{li2023steganerf} or NeRF Steganography \cite{dong2023steganography} explored embedding a 2D image into the NeRF model. However, these methods are not only limited in capacity to 2D data, but also \wjx{frequently} require an additional external decoder network to extract the hidden information, which makes the existence of steganography 
\jwx{easily detectable}
in practice. 

Recently, GS-Hider \cite{zhang2024gs} has made significant progress. It is based on the 3DGS model \cite{kerbl20233d} and successfully hides another complete 3D scene by adding extra Spherical Harmonics (SH) parameters to each Gaussian and an auxiliary decoder. Although GS-Hider achieves high-capacity hiding, its method introduces significant parameter overhead and changes the model's base architecture. This architectural modification and increase in parameter count make the model structurally detectable ($i.e.$, detectable by its parameter count or pipeline structure), violating the core goal of steganography.

\subsection{Other NeRF Information Hiding Methods}

In addition to the watermarking and steganography methods discussed above, IPA-NeRF \cite{jiang2024ipa} and GaussTrap \cite{hong2025gausstrap} utilize a secret viewpoint as a backdoor trigger to hide information within the model. 
\wjx{Specifically,} when the model renders from ordinary viewpoints, it displays the original cover scene; however, when rendering from a preset secret viewpoint, it reveals the hidden backdoor image.

While this approach is conceptually \jwx{elegant} and avoids the overhead of an extra decoder, it still suffers from limitations. First, its hiding capacity is typically limited to a single 2D image, making it incapable of hiding an entire 3D scene. Second, this method may actually cause local damage to the 3D model's integrity. This is fundamentally different from our StegoNGP, which uses a cryptographic hash key that is completely decoupled from the model's rendering.
\section{Method}

\subsection{Preliminary}
\wjx{In this section, we} first review the basic mathematical definitions for NeRF rendering and the Instant-NGP multi-resolution hash feature embedding. These definitions establish the groundwork for the subsequent problem definition and introducing \wjx{the proposed StegoNGP.}

\paragraph{Neural Radiance Fields (NeRF).} For a NeRF model $F:(\mathbf{x}, \mathbf{d})\to(c, \tau)$, the input is a five-tuple, the coordinates of the sampling point $\mathbf{x} \in \mathbb{R}^3$ and the direction of the sampling ray $\mathbf{d}\in\mathbb{R}^3$, the output is an RGB color $c\in[0,1]^3$ and a volume density $\tau \in \mathbb{R}^+$. Each pixel in the target image represents one ray $\mathbf{x} = \mathbf{r}(t) := \mathbf{o} + t\mathbf{d}$ in space, which emanates from the camera centre $\mathbf{o}$ towards the ray direction $\mathbf{d}$, where $t$ is the ray depth. Along the ray direction, the output from discrete $n$ sampling points is integrated to obtain the predicted color value $\hat{C}$ for a pixel, as follows:
\begin{align}
\hat{C}(\mathbf{r}, F) & := \sum^n_{i=1} T(t_i)\cdot \alpha(\tau(t_i)\cdot\delta_i)\cdot c(t_i), \\
 T(t_i) &:= \exp (\ - \sum^{i-1}_{j=1} \tau(t_j)\cdot\delta_j ),\ 
\end{align}
where $\alpha(x) := 1-\exp(-x)$, $\delta_i:=t_{i+1}-t_i$ denotes the distance between two sampling points, $c(t_i)$ and $\tau(t_i)$ represent the color and density at $\mathbf{r}(t_i)$.

\paragraph{Instant Neural Graphics Primitives (Instant‑NGP).} Instant-NGP achieves training and rendering speeds several times faster than the original NeRF by using multi-resolution 3D hash feature embeddings and a shallow MLP to represent the 3D scene. Specifically, Instant-NGP partitions the space into grids at multiple resolution levels. The resolution $N_l$ for each level $l$ scales geometrically between the coarsest level $N_{min}$ and finest level $N_{max}$, as follows:

\begin{align}
N_l:=&[N_{min} \cdot z^l], \\
z:=&\exp \left( \frac{\ln N_{max} - \ln N_{min}}{L - 1} \right).
\end{align}

For a given sampling position $\mathbf{x}$, the Instant-NGP model $\mathcal{M}:(\mathcal{T}(h(\mathbf{x})), \mathbf{d}) \to (c, \tau)$ \wjx{utilizes} a four-layer MLP to map an encoding to an RGB color $c$ and a volume density $\tau$. This encoding is the concatenation of feature vectors from multiple resolutions, obtained via the hash table lookup $\mathcal{T}(\cdot)$. The hash table indices are computed from the sampling position $\mathbf{x}$ by the hash function $h(\mathbf{x})$, as follows:
\begin{align}
h(\mathbf{x}):= \left(\bigoplus^{d}_{i=1}x_i\pi_i \right)\mod T,
\end{align}
where $\oplus$ denotes the bit-wise XOR operation and $\pi_i$ are unique large prime numbers, $d$ represents the dimension of the input coordinate $\mathbf{x}$, and $T$ is the max entries per level ($i.e.$, the hash table size).

\subsection{Problem Formulation}

Consider a cryptographic steganography model $\mathcal{H}$ designed for dual functionality. It encrypts and embeds a hidden scene $B$, which can be decoded given an encryption key $\mathcal{K}$, while rendered using its default pipeline $\mathcal{M}$ without the key, it presents the original cover scene $S$. Then, the aim of the cryptographic steganography is as follows:
\begin{align}
\begin{split}
 \min_{\mathcal{H}}\forall_{v \in V} ( \| \mathcal{I} (\mathcal{H} (\mathcal{K}), v) - & b_{v} \|^2_2 \\ 
 + & \| \mathcal{I}(\mathcal{M}, v) - s_{v} \|^2_2 ) ,
\end{split} \label{eq:eqcrypt_stega}
\end{align}
where $V$ is the set of observed viewpoints, $\mathcal{I}$ represents the compositing function, which integrates the colors and volume densities obtained from sampling points along the ray to compose the predicted image $\hat{I}_v$ for viewpoint $v \in V$, while $b_{v} \in B$ and $s_{v} \in S$ are the ground truths image of the hidden scene $B$ and the cover scene $S$ from viewpoint $v$, respectively. Let $\mathcal{M}$ be the standard Instant-NGP model, which is defined as the degeneration \jwx{form} of $\mathcal{H}$ when using the default large prime number sequence $\Pi$ ($i.e.$, $\mathcal{M} \equiv \mathcal{H} (\Pi)$), occurring in the absence of the decryption key.

\subsection{Cryptographic Steganography} \label{subsec:cs}

For $d=3$, the original Instant-NGP model \wjx{utilize} the default prime sequence $\Pi \equiv \{1, 2654435761, 805459861\}$. We define our keyed hash indexing function $\psi(\mathbf{x},\mathcal{K})$ by generalizing $h(\mathbf{x})$ to replace $\Pi$ with a secret key $\mathcal{K}\equiv\{k_1, k_2, k_3\}$, as follows:

\begin{align}
\psi(\mathbf{x},\mathcal{K}):= \left(\bigoplus^{d}_{i=1}x_i k_i \right) \!\!\! \mod T.
\end{align}

Moreover, by switching the encryption key between the default sequence and a given secret sequence (e.g., $\mathcal{K} = \{54857899, 1455645677, 5487678709\}$), we can embed two scenes into the model simultaneously, the cover scene $\mathcal{S}$ and the hidden scene $\mathcal{B}$. The StegoNGP $\mathcal{H}$, which is optimized to solve the problem in Equation \ref{eq:eqcrypt_stega}, is defined as follows:

\begin{align}
\mathcal{H}(\mathcal{K},\mathcal{S},\mathcal{B}) := \begin{cases}
\mathcal{M}(\mathcal{T}(\psi(\mathbf{x},\mathcal{K})), \mathbf{d}), &\!\!\! \text{for hidden scene } \mathcal{B} \\
\mathcal{M}(\mathcal{T}(h(\mathbf{x})), \mathbf{d}), & \!\!\! \text{for cover scene } \mathcal{S}.
\end{cases}
\end{align}

As depicted in Algorithm~\ref{alg:cs_algo}, each training iteration involves randomly selecting either the cover scene $S$ or the hidden scene $B$ to fit. The default large prime number sequence $\Pi$ is used for the cover scene, while the encryption key $\mathcal{K}$ is used for the hidden scene. The MSE loss $\mathcal{L}_{mse}$ and the weight sparsity loss $\mathcal{L}_{spar}$ are calculated and used to train the StegoNGP $\mathcal{H}$.

\begin{algorithm}[h]
	\caption{Cryptographic steganography}
	\label{alg:cs_algo}

    \textbf{Input:} $S$: cover scene, $B$: hidden scene \\
    \textbf{Input:} $\Pi$: default large prime numbers list\\
    \textbf{Input:} $\mathcal{K}$: steganographic encryption key \\
    \textbf{Input:} $\mathcal{M}$: initialization Instant-NGP model \\  
    \textbf{Input:} $\beta$: loss weight, $\eta$: learning rate \\ 
    \textbf{Input:} $K$: maximum iterations \\
    \textbf{Output:} $\mathcal{H}$: StegoNGP that includes both cover and hidden scenes \\
    \begin{algorithmic}[1]
        \State $\mathcal{H} \gets \mathcal{M}$
        \While{$i < K$}	
            \State $\mathcal{S}^* \gets random.choice([S,B])$
            \If {$\mathcal{S}^*$ is $S$}
                \State $\mathcal{K}^* \gets \Pi$ 
            \Else 
                \State $\mathcal{K}^* \gets \mathcal{K}$
            \EndIf
            \State{$I_v, v \gets random.choice(\mathcal{S}^*)$}
            \State $ \hat I_v \gets \mathcal{I}(\mathcal{H}(\mathcal{T}(\psi(\mathbf{x},\mathcal{K}^*))), v) $ 
            \State $\mathcal{L} \gets \mathcal{L}_{mse}(\hat I_v, I_v) + \beta \cdot \mathcal{L}_{spar}(\mathcal{H})$ 
            \State $\mathcal{H} \gets \mathcal{H} + \eta \cdot \nabla_{\mathcal{H}}\mathcal{L}$ 
            
        \EndWhile
    \end{algorithmic} 
\end{algorithm}

\begin{figure*}[bt]
  \includegraphics[width=1.0\textwidth] {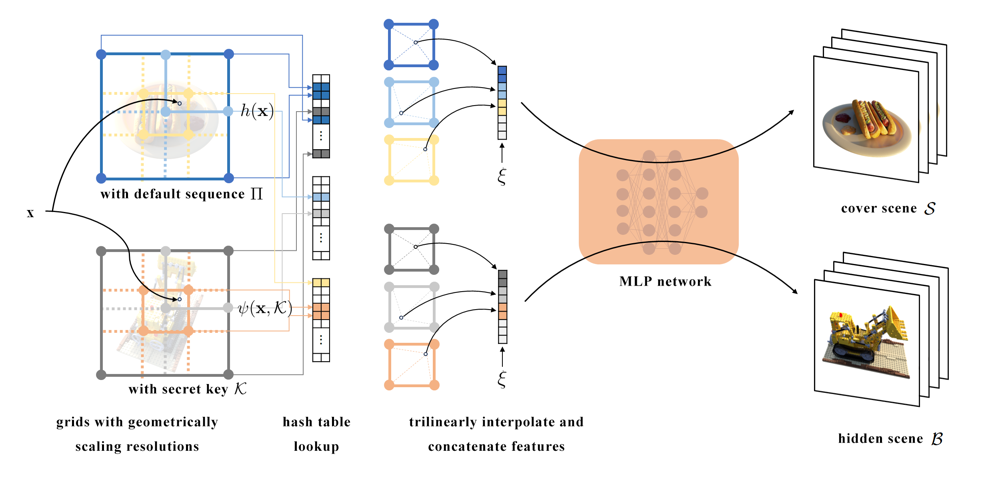}
  \caption{StegoNGP framework for dual-scene representation.}
  \label{fig:framework}
\end{figure*}

Figure \ref{fig:framework} intuitively shows our proposed StegoNGP framework, where the MLP network renders a scene using features from a shared hash table $\mathcal{T}$ and an auxiliary input $\xi$ (such as the encoded view direction). The core of our framework is the key-controlled hash function that queries table $\mathcal{T}$. When the default large prime number sequence $\Pi$ is used through the standard hash function $h(\mathbf{x})$, the framework reconstructs or renders the cover scene $\mathcal{S}$. Conversely, when the secret key $\mathcal{K}$ is provided to the hash function $\psi(\mathbf{x}, \mathcal{K})$, the function computes different indices, causing the exact same network parameters to reconstruct or render the hidden scene $\mathcal{B}$. 

\subsection{Multi-Key Level-Assignment Scheme}\label{subsec:multi_key}

While the method described in Section \ref{subsec:cs} already provides robust cryptographic security, its security guarantee can be further enhanced for applications requiring an even higher level of protection. We introduce this enhanced configuration as Multi-Key StegoNGP, which uses the Multi-Key Level-Assignment scheme. In the standard Instant-NGP, there are $L=16$ resolution levels, and the default hash function $h(\mathbf{x})$ \wjx{utilizes} the same large prime sequence $\Pi$ uniformly across all levels. Our basic StegoNGP (described in Section \ref{subsec:cs}) replaces this with a single secret key $\mathcal{K}$, which is also applied uniformly to all $L$ levels. We can strengthen this by defining a key set $\mathbf{K}$ containing $m$ distinct secret keys:
\begin{align}
\mathbf{K} = \{\mathcal{K}_1, \mathcal{K}_2, \dots, \mathcal{K}_m\}, \quad \text{where } 1 \le m \le L,
\end{align}
instead of applying one key to all levels, we assign these $m$ keys to the $L$ levels in a cyclic fashion. Let $l \in \{1, \dots, L\}$ be the level index. The specific key $\mathcal{K}^{(l)}$ assigned to level $l$ is determined by its index modulo $m$:
\begin{align}
\mathcal{K}^{(l)} = \mathcal{K}_{j}, \quad \text{where} \quad j = ((l-1) \!\!\! \mod m) + 1.
\end{align}

This mapping allows for flexible and enhanced security configurations:

\begin{itemize}
\item \textbf{When} $\mathbf{m=1}$: The key set is $\mathbf{K} = \{\mathcal{K}_1\}$, this 
\jwx{reduces} to our basic StegoNGP.

\item \textbf{When} $\mathbf{m=2}$: The keys $\mathbf{K} = \{\mathcal{K}_1, \mathcal{K}_2\}$ are applied to odd and even levels, respectively.

\item \textbf{When} $\mathbf{m=L=16}$: This provides the maximum security configuration, where each of the 16 levels employs a unique \wjx{and} independent secret key.
\end{itemize}

This multi-key approach significantly increases the cryptographic key space. An attacker can no longer compromise the entire hidden scene by discovering a single key $\mathcal{K}$. They 
\wjx{are required to} discover all $m$ keys and the correct assignment order, making the model far more robust against cryptographic analysis and brute-force attacks.
\section{Experiments}

In this section, we 
\wjx{first detail} the basic setting for the experiments. \wjx{Then, we }
demonstrate the effectiveness of our method on synthetic scenes \wjx{by} comparing it against the baseline Instant-NGP model. Next, we analyze the Key Sensitivity and Plausible Deniability. Following this, we provide a comparison with existing 3D steganography methods without encryption. Finally, we explore its generalization capability on real-world scenes and conduct a thorough ablation study on the multi-key scheme.

\subsection{Experiments Setting}

\paragraph{Dataset.}
\wjx{All the experiments are conducted on the Blender Synthetic dataset and the Mip-Nerf-360 dataset. }
The Blender Synthetic dataset, introduced in the original NeRF paper \cite{mildenhall2021nerf}, contains eight objects. Each object consists of $400$ images generated from different viewpoints sampled on the upper hemisphere with a resolution of $800 \times 800$ pixels, which are split into $100$ images for training, $200$ for testing, and $100$ for validation. In addition, we synthesized two additional 3D scenes (Google and Twitter) in the same format using Blender to serve as our hidden scenes. The Mip-Nerf-360 dataset \cite{barron2022mip} includes nine real-world captured scenes. To better reflect real-world application scenarios, it features an uneven distribution of data across its training, testing, and validation sets. For this dataset, we pre-process its images and generate camera poses via COLMAP.

\paragraph{Model.}
Our baseline model is the standard Instant-NGP \cite{muller2022instant}, trained for $50,000$ iterations. Our StegoNGP shares this architecture and total iteration count ($50,000$). At each iteration, it randomly targets either the cover scene $\mathcal{S}$ (using default $\Pi$) or the hidden scene $\mathcal{B}$ (using key $\mathcal{K}$ or Multi-keys $\mathbf{K}$). We \wjx{further compare our StegoNGP against GS-Hider \cite{zhang2024gs} by using its default configuration and the same $50,000$ iteration limit.}


\subsection{Performance on Synthetic Scene}\label{sec:sec_Synthetic_Scene}

We provide an evaluation of the steganographic effectiveness of our StegoNGP framework on the Blender Synthetic dataset. We compare our method against the Instant-NGP (baseline) to demonstrate its capability to hide a complete 3D scene while maintaining high reconstruction fidelity.

\newcommand{\figt}[2][1]{\includegraphics[width=#1\linewidth]{images/main_experiment/#2}}
\newcommand{\sizeM}{1}
\begin{figure*}[t]
\scriptsize
\centering
\setlength{\tabcolsep}{1pt}
\begin{tabular}{m{2cm}m{2cm}m{2cm}m{2cm}m{2cm}m{2cm}m{2cm}m{2cm}}

     Chair (Twitter) 
     &  \figt[\sizeM]{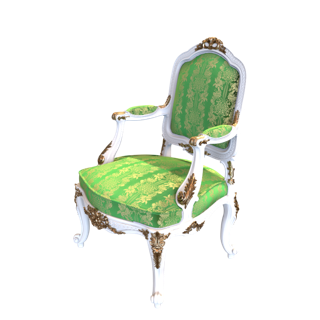}
     &  \figt[\sizeM]{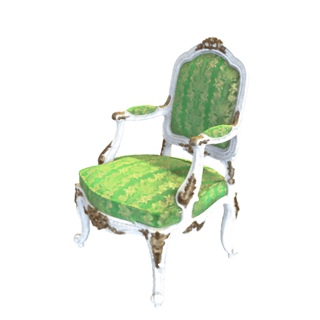}
     &  \figt[\sizeM]{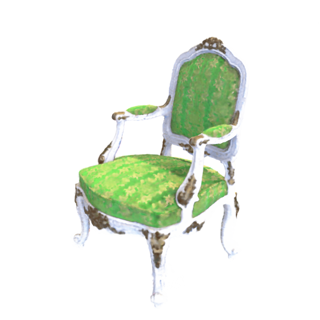}
     &  \figt[\sizeM]{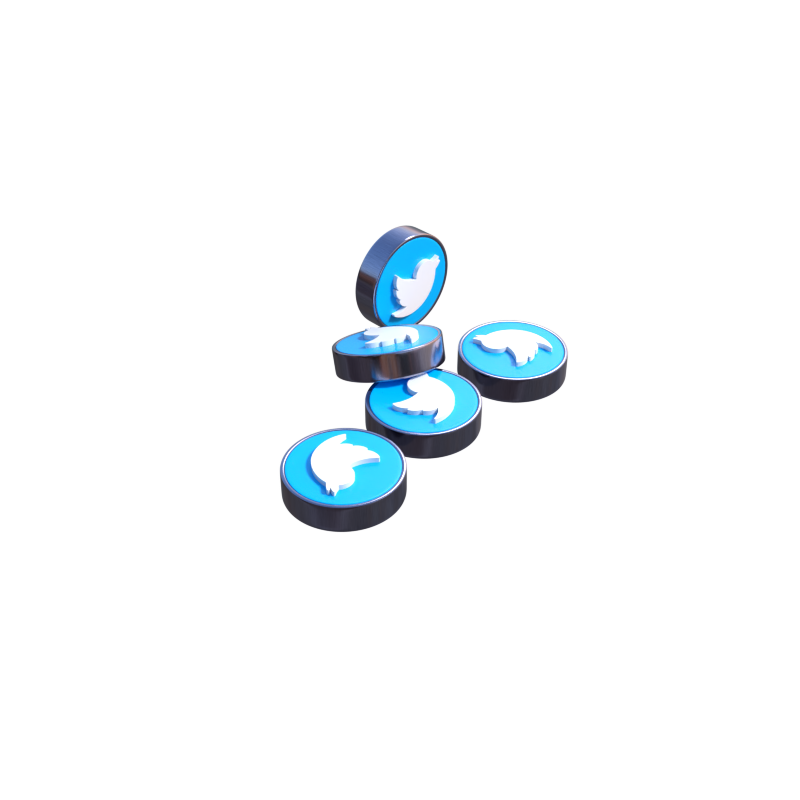}
     &  \figt[\sizeM]{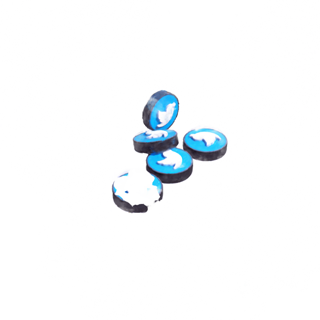}
     &  \figt[\sizeM]{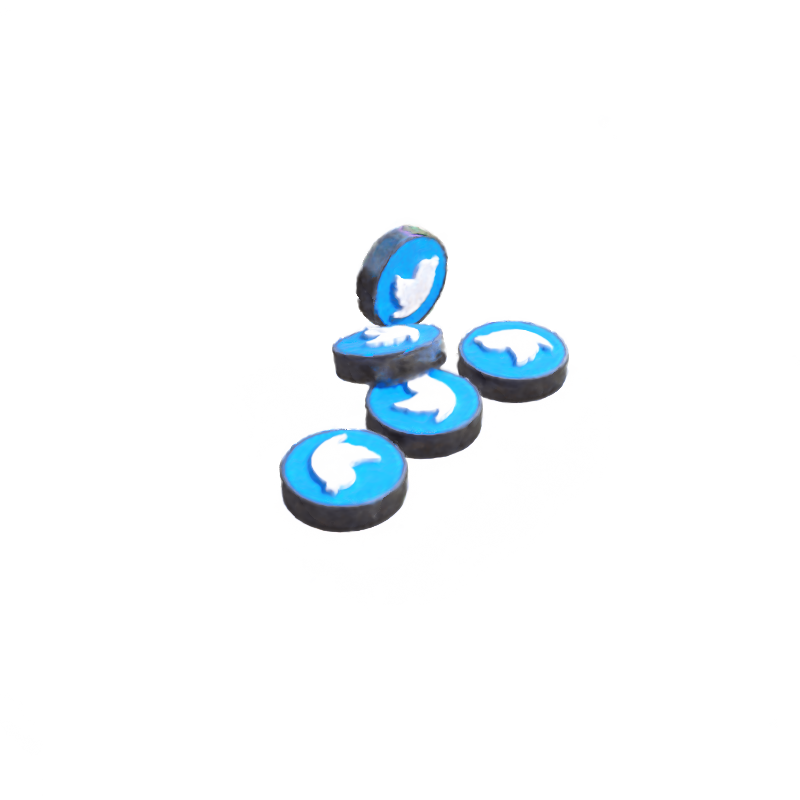}
     &  \figt[\sizeM]{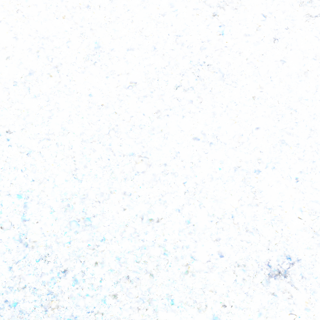} \\

     Ficus (Twitter) 
     &  \figt[\sizeM]{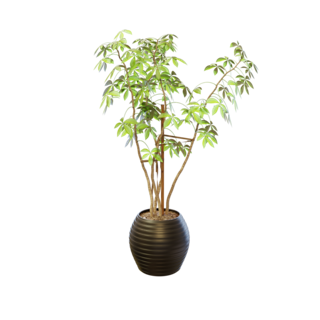}
     &  \figt[\sizeM]{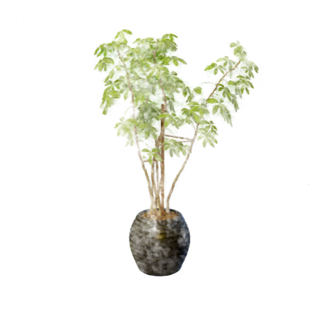}
     &  \figt[\sizeM]{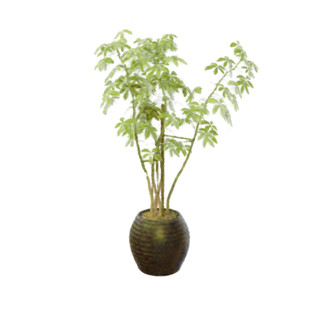}
     &  \figt[\sizeM]{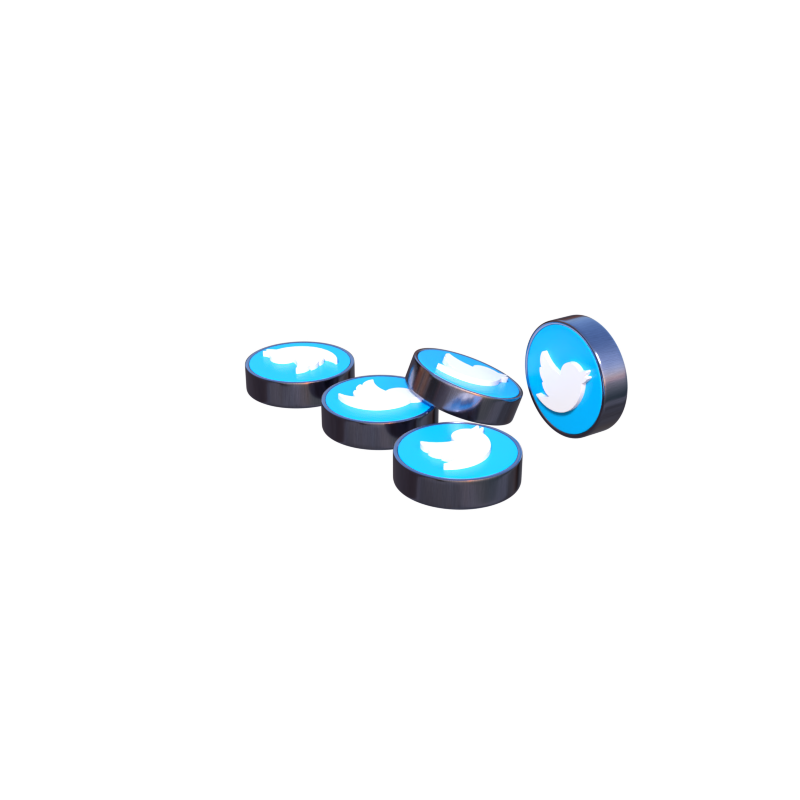}
     &  \figt[\sizeM]{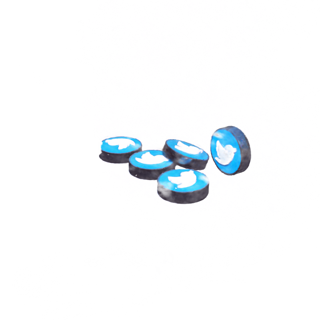}
     &  \figt[\sizeM]{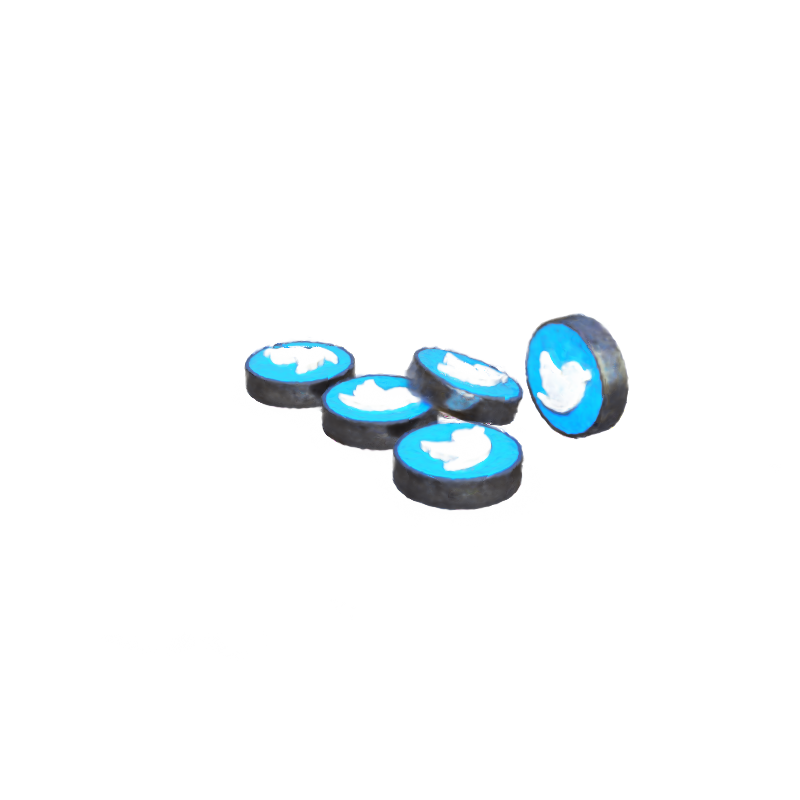} 
     &  \figt[\sizeM]{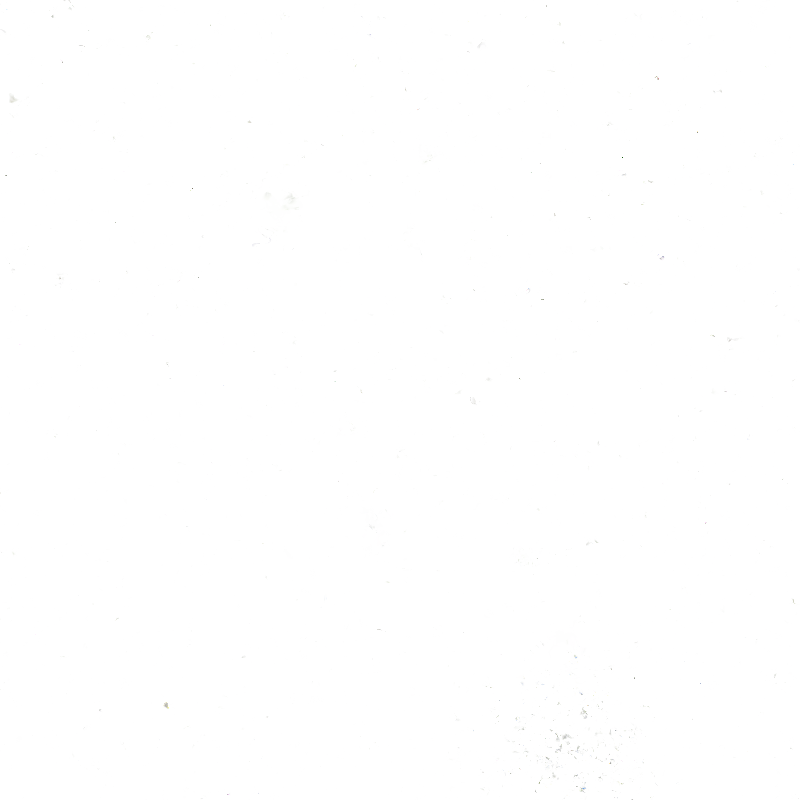} \\  

     Hotdog (Google) 
     &  \figt[\sizeM]{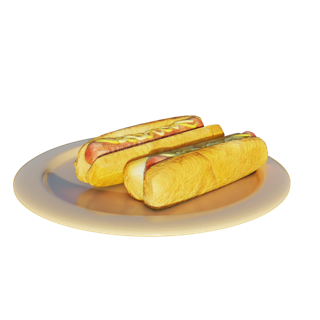}
     &  \figt[\sizeM]{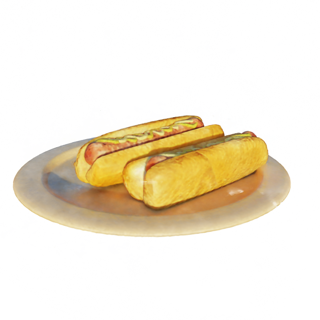}
     &  \figt[\sizeM]{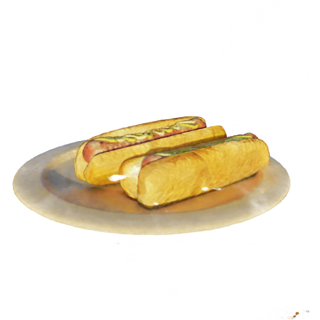}
     &  \figt[\sizeM]{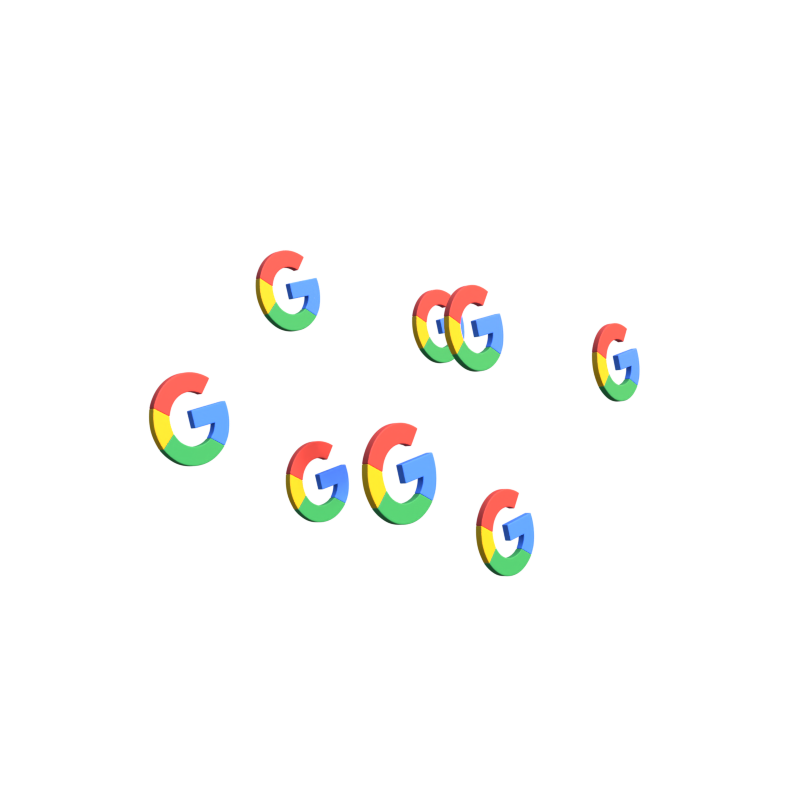}
     &  \figt[\sizeM]{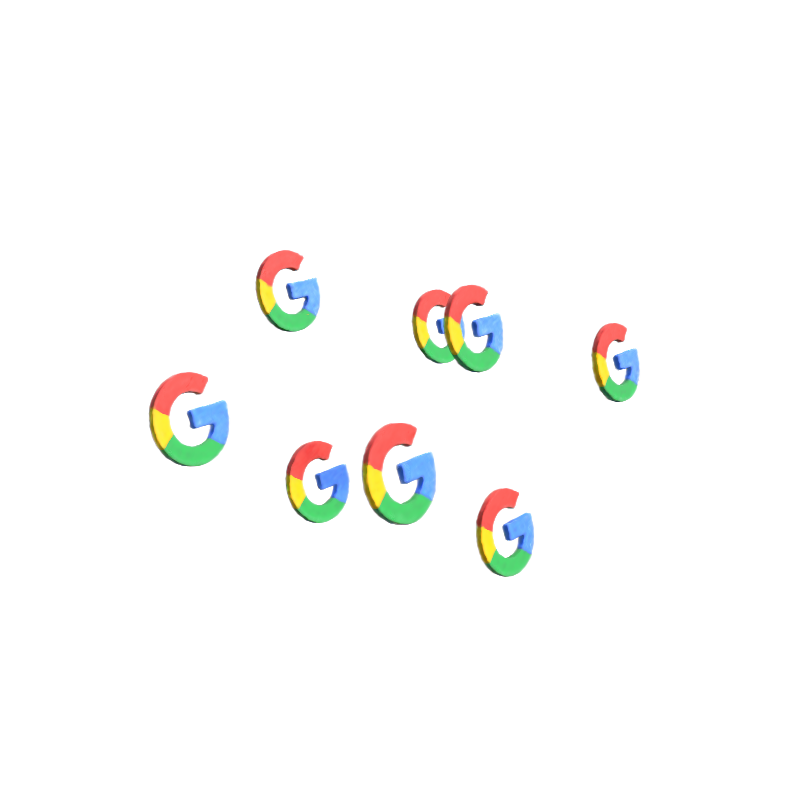}
     &  \figt[\sizeM]{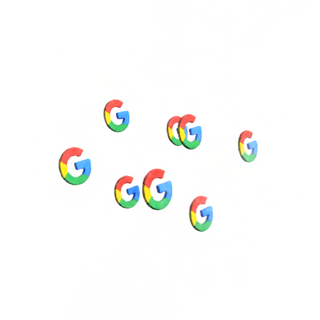}
     &  \figt[\sizeM]{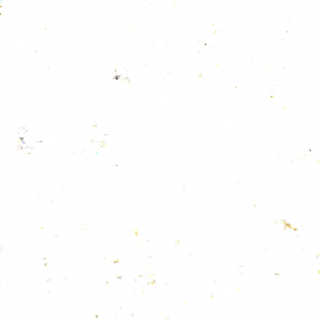} \\
     
     Ship (Google)
     &  \figt[\sizeM]{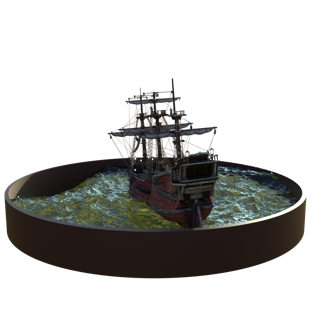}
     &  \figt[\sizeM]{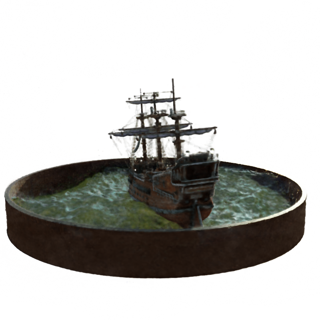}
     &  \figt[\sizeM]{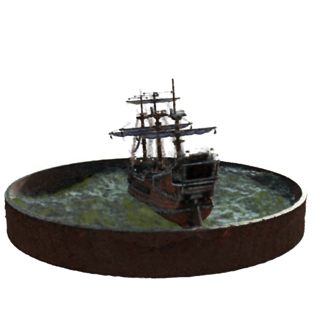}
     &  \figt[\sizeM]{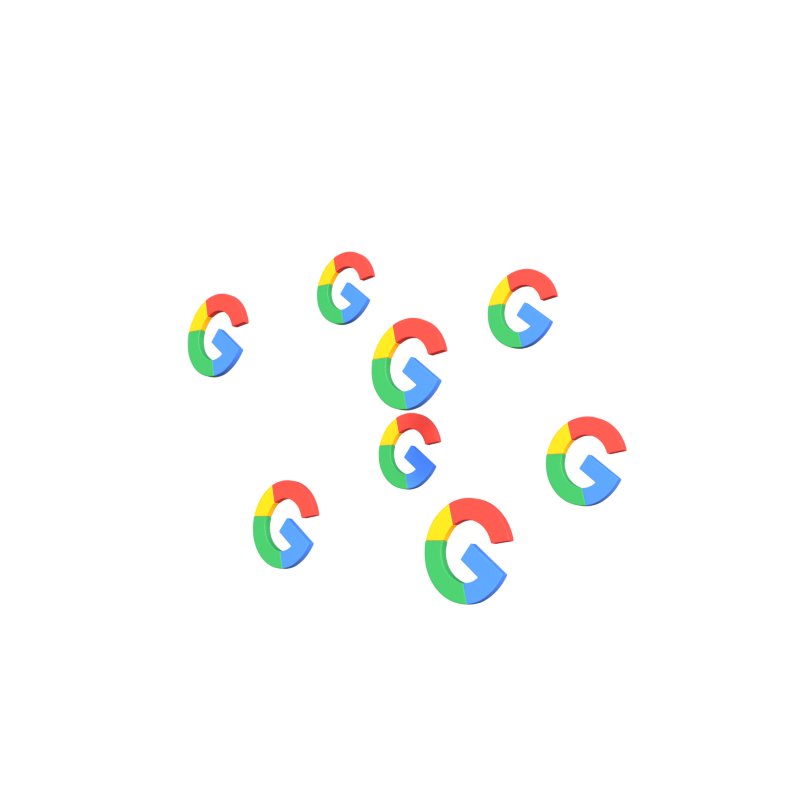}
     &  \figt[\sizeM]{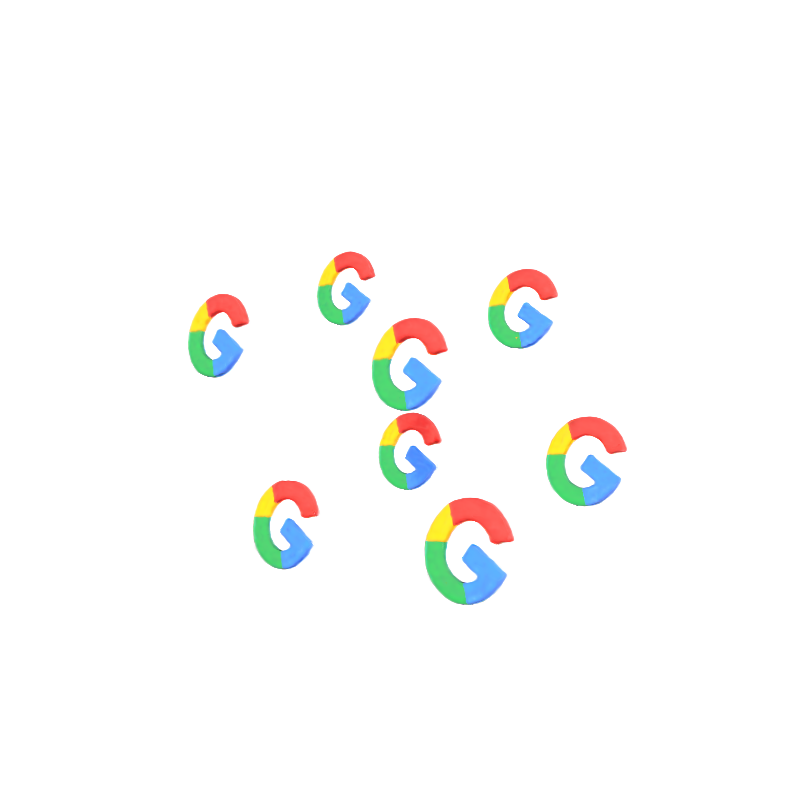}
     &  \figt[\sizeM]{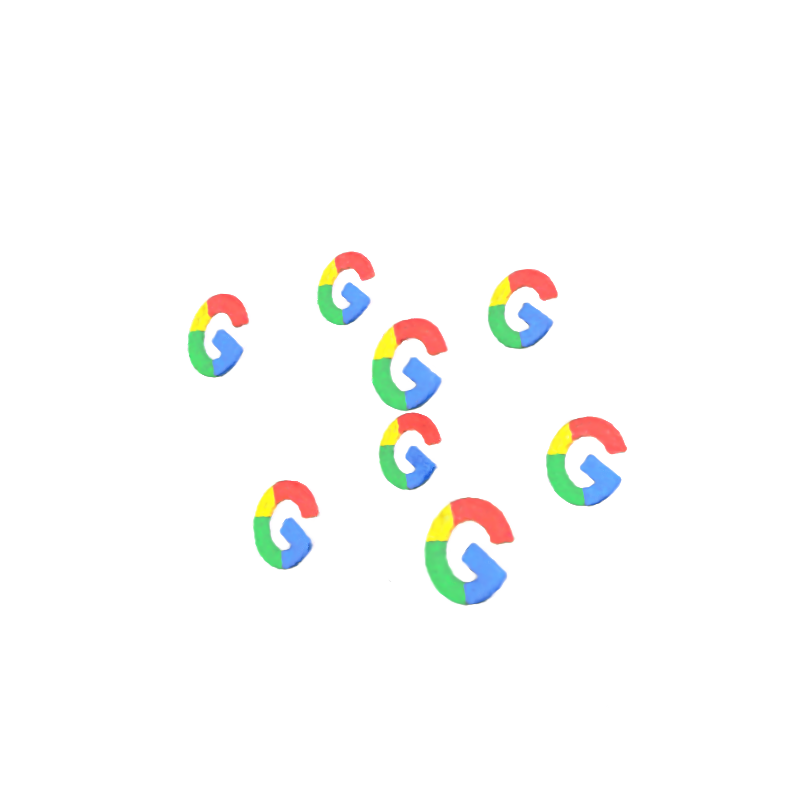} 
     &  \figt[\sizeM]{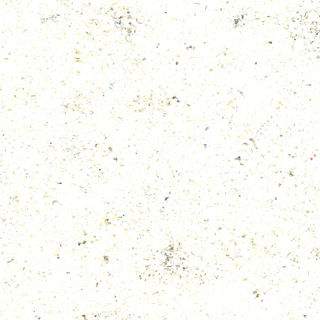} \\  
     
    \hspace{0.1cm} cover scene 
    & \multirow{2}{*}{\hspace{0.7cm}GT ($S$)} 
    & \multirow{2}{*}{\hspace{0.6cm}INGP ($S$)}
    & \multirow{2}{*}{\hspace{0.3cm}\textbf{StegoNGP} ($\Pi$)}
    & \multirow{2}{*}{\hspace{0.7cm}GT ($\mathcal{B}$)} 
    & \multirow{2}{*}{\hspace{0.6cm}INGP ($\mathcal{B}$)}
    & \multirow{2}{*}{\hspace{0.3cm}\textbf{StegoNGP} ($\mathcal{K}$)} 
    & \multirow{2}{*}{\hspace{0.3cm}\textbf{StegoNGP} ($\mathcal{O}$)} 
    \\
    
    (hidden scene) & & & & & & \\
    
   \end{tabular}
   \caption{Visualization comparing the Ground Truth (GT) images of the cover scene $\mathcal{S}$ and hidden scene $\mathcal{B}$ against the rendered outputs from the baseline Instant-NGP (INGP) and our StegoNGP. Our model renders the scene $\mathcal{S}$ using the default key $\Pi$ and reveals the scene $\mathcal{B}$ using the secret key $\mathcal{K}$. It also shows the blank or noise 
   \jwx{outputs}
   from an unrelated key $\mathcal{O}$.}
    \label{fig:visual_main_experiment}
\end{figure*}

For this experiment, we utilized the eight standard objects from the Blender dataset as cover scenes ($\mathcal{S}$) and our two custom-synthesized scenes (Google and Twitter) as hidden scenes ($\mathcal{B}$). We formed eight 
\jwx{experiment}
pairs by assigning each cover scene to one of the two hidden scenes. Specifically, four scenes (Lego, Mic, Chair, and Ficus) were paired with Twitter, while the other four (Hotdog, Drums, Ship, and Materials) were paired with Google. The Instant-NGP (baseline) models were trained independently on each of the 10 scenes to establish the upper bound for reconstruction quality. Our StegoNGP was trained on all eight pairs, and the quantitative results in Table \ref{table:main_experiment} represent the average performance across these eight experiments.

\begin{table}[htp]
\centering
\begin{tabular}{ccccc}
\toprule
Model & Scene & PSNR$\uparrow$ & SSIM$\uparrow$ & LPIPS$\downarrow$ \\
\midrule
Instant-NGP & $\mathcal{S}$ & 26.508 & \textbf{0.913} & \textbf{0.078} \\
(baseline) & $\mathcal{B}$ & 32.436 & 0.985 & 0.014 \\
\midrule
\textbf{StegoNGP} & $\mathcal{S}$ & \textbf{26.581} & 0.910 & 0.083 \\ 
\textbf{(Ours)} & $\mathcal{B}$ & \textbf{34.048} & \textbf{0.988} & \textbf{0.007} \\
\bottomrule
\end{tabular}
\caption{Quantitative comparison of reconstruction quality (PSNR, SSIM, and LPIPS) between StegoNGP (ours) and Instant-NGP (baseline) on the Blender Synthetic dataset. The baseline method reconstructs each scene independently, whereas our StegoNGP method reconstructs the cover scene $\mathcal{S}$ with one embedded hidden scene $\mathcal{B}$.}
\label{table:main_experiment}
\end{table}

As shown in Table \ref{table:main_experiment}, our StegoNGP successfully reconstructs both the cover scene $\mathcal{S}$ and the hidden scene $\mathcal{B}$ with high fidelity. The quantitative metrics are highly comparable to (and in some cases, slightly surpass) the Instant-NGP baseline. This indicates that our joint-training process does not introduce any significant quality degradation. The slight improvement observed in this dataset might be attributed to a regularization effect.

Figure \ref{fig:visual_main_experiment} shows that the rendered images from StegoNGP closely match the baseline renders. This is the key requirement for steganographic imperceptibility, as the cover scene $\mathcal{S}$ does not 
\jwx{exhibit} any visual artifacts that would distinguish it from a standard Instant-NGP model. \wjx{As shown in Figure~\ref{fig:visual_main_experiment}, the model's cryptographic security is demonstrated by the incoherent noise produced by an unrelated key $\mathcal{O}$.}

\begin{table*}[htbp]
\centering
\begin{tabular}{cccc|ccc|cc}
\toprule
Scene & \multicolumn{3}{c|}{cover scene $\mathcal{S}$} & \multicolumn{3}{c|}{hidden scene $\mathcal{B}$} & additional parameters & structural \\
Model & PSNR$\uparrow$ & SSIM$\uparrow$ & LPIPS$\downarrow$ & PSNR$\uparrow$ & SSIM$\uparrow$ & LPIPS$\downarrow$ & Model Size $\downarrow$ & detectability\\
\midrule
GS-Hider \cite{zhang2024gs} & \textbf{36.194} & \textbf{0.991} & \textbf{0.006} & \textbf{29.625} & \textbf{0.962} & \textbf{0.030} & + 20.54 MB & + \\
\textbf{StegoNGP (ours)} & 29.678 & 0.946 & 0.056 & 26.248 & 0.897 & 0.094 & \textbf{+ 0 MB} & \textbf{-} \\  
\bottomrule
\end{tabular}
\caption{Comparison of StegoNGP (ours) with the GS-Hider \cite{zhang2024gs} method across reconstruction fidelity (PSNR, SSIM and LPIPS), additional model size (MB), and structural detectability.}
\label{table:compared_with_gs_hider}
\end{table*}

\subsection{Key Sensitivity and Plausible Deniability}\label{sec:sec_Key_Sensitivity}

\wjx{Besides} reconstruction fidelity, the core contribution of StegoNGP 
\wjx{depends on} 
its cryptographic security. In this subsection, we analyze the model's robustness against key attacks and its capacity for plausible deniability.

\begin{figure}[t]
\scriptsize
\centering
\setlength{\tabcolsep}{1pt}
\begin{tabular}{m{2.0cm}m{2.0cm}m{2.0cm}m{2.0cm}}
        $\{k_1, k_2, k_3\}$
     &  \includegraphics[width=1\linewidth]{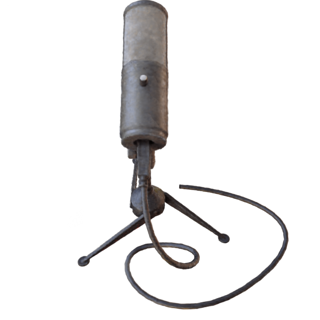}
     &  \includegraphics[width=1\linewidth]{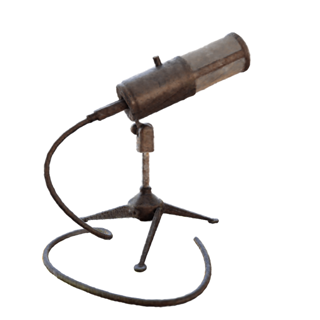}
     &  \includegraphics[width=1\linewidth]{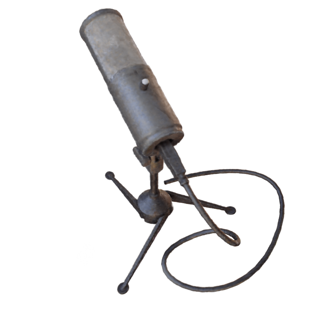}\\

        $\{\pi_1, k_2, k_3\}$
     &  \includegraphics[width=1\linewidth]{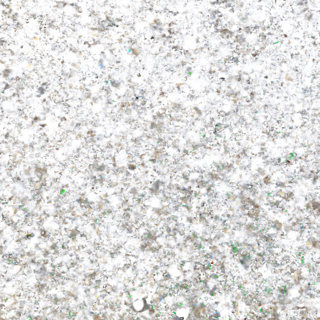}
     &  \includegraphics[width=1\linewidth]{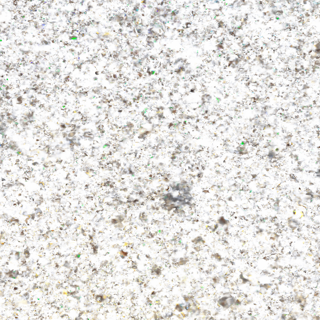}
     &  \includegraphics[width=1\linewidth]{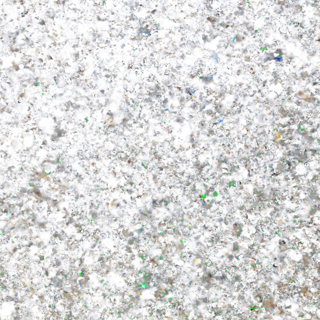}\\

        $\{\pi_1, \pi_2, k_3\}$
     &  \includegraphics[width=1\linewidth]{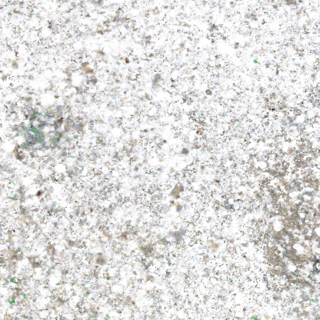}
     &  \includegraphics[width=1\linewidth]{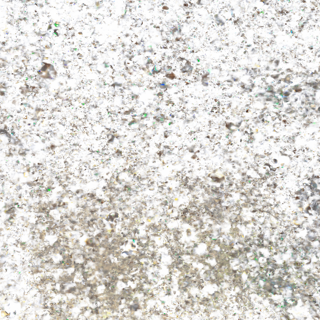}
     &  \includegraphics[width=1\linewidth]{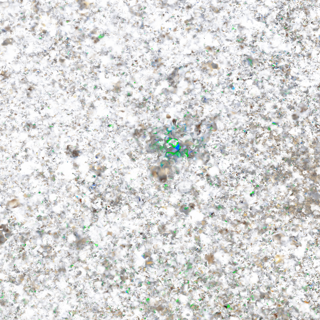}\\

        $\{\pi_1, \pi_2, \pi_3\}$
     &  \includegraphics[width=1\linewidth]{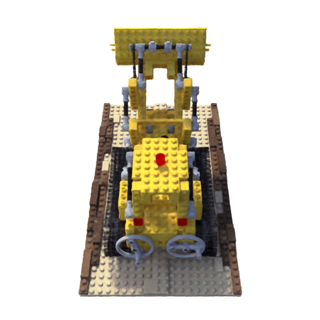}
     &  \includegraphics[width=1\linewidth]{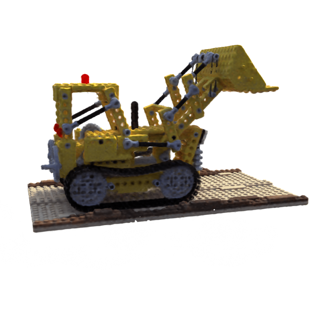}
     &  \includegraphics[width=1\linewidth]{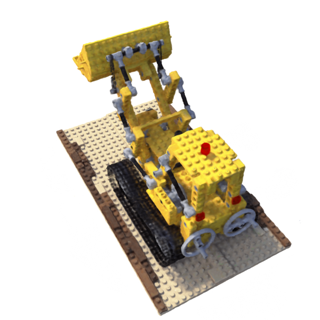}\\

       $\{o_1, o_2, o_3\}$
     &  \includegraphics[width=1\linewidth]{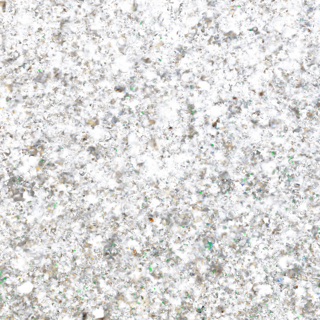}
     &  \includegraphics[width=1\linewidth]{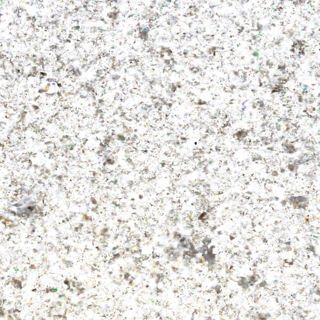}
     &  \includegraphics[width=1\linewidth]{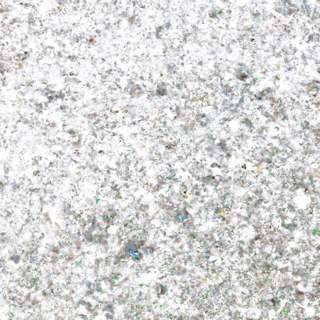}\\

        \hspace{0.3cm}hash key
     &  \multicolumn{3}{c}{rendered image (test set)}\\
     
\end{tabular}
   \caption{Demonstration of key sensitivity and security in the StegoNGP. The model renders the cover scene (Lego) and hidden scene (Mic) only when the precise default key $\{\pi_1, \pi_2, \pi_3\}$ or secret key $\{k_1, k_2, k_3\}$ is provided. Partially correct (mixed) keys or other incorrect keys fail to decode any coherent information, resulting in noise.}
    \label{fig:visual_different_keys}
\end{figure}

\wjx{To verify the effectiveness of the proposed model,} we 
\wjx{leveraged} 
a 
trained StegoNGP (Lego as $\mathcal{S}$ and Mic as $\mathcal{B}$) and fed it four different types of keys, the default key $\Pi$, the correct secret key $\mathcal{K}$, partially correct (mixed) keys, where some primes are from $\Pi$ and some from $\mathcal{K}$, and an entirely incorrect key $\mathcal{O} \equiv\{o_1, o_2, o_3\}$. The results 
\wjx{can be found in} 
Figure \ref{fig:visual_different_keys}.


As demonstrated in Figure \ref{fig:visual_different_keys}, the StegoNGP exhibits two critical security features. First, when provided with the default key $\Pi$, the model renders the cover scene $\mathcal{S}$ perfectly, establishing plausible deniability as it appears identical to a standard, non-steganographic model. Second, the model shows high key sensitivity. The hidden scene $\mathcal{B}$ can only be recovered with the exact secret key $\mathcal{K}$. Any attempt with partially correct (mixed) keys or other incorrect keys fails, resulting in incoherent noise. This confirms that the security relies on the entire secret key, making it robust against brute-force or partial-discovery attacks.

\subsection{Comparison with Existing Methods} \label{sec:sec_Existing_Methods}

We provide a comparison between our StegoNGP and GS-Hider \cite{zhang2024gs}, an existing steganography method based on 3DGS. To accommodate GS-Hider's requirement for coupled camera poses (where cover and hidden scenes must share identical poses), we generated a specialized Blender dataset for this comparison. We re-synthesized all 8 scenes with aligned camera matrices (100 train, 100 test, 100 validation views each) and created four scene pairs (Hotdog-Lego, Mic-Drums, Chair-Ship, and Ficus-Materials). The experimental results presented below are the average of these four pairs. Unlike our cryptographic, zero-overhead approach, GS-Hider alters the model architecture and relies on additional neural network decoders to separate the cover and hidden scenes. We analyze the trade-offs between these two strategies in terms of reconstruction fidelity, model size overhead, and structural detectability.

As shown in Table \ref{table:compared_with_gs_hider}, while GS-Hider leads in scene reconstruction fidelity, this advantage comes at the cost of significant model overhead. The method introduces an average of 20.54MB of extra parameters and alters the model architecture, which makes it structurally detectable. 
\jwx{Even without a complex classifier, an external observer} 
can accurately identify the GS-Hider model simply by checking its file size or parameter count. This clearly 
\jwx{undermines the fundamental goal of} steganography. In contrast, our StegoNGP achieves zero parameter overhead, making it indistinguishable from a standard Instant-NGP model in both architecture and size.

\subsection{Performance on Real World Scene} \label{sec:sec_Real_World}

\begin{figure*}[htp]
\scriptsize
\centering
\setlength{\tabcolsep}{1pt}
\begin{tabular}{m{1.0cm}m{1.95cm}m{1.95cm}m{1.95cm}m{1.95cm}m{1.95cm}m{1.95cm}m{1.95cm}m{1.95cm}}

     {Bonsai (Kitchen) } 
     &  \includegraphics[width=1\linewidth]{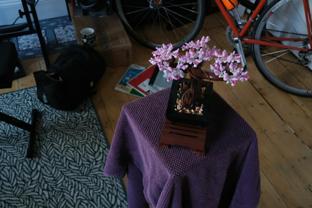}
     &  \includegraphics[width=1\linewidth]{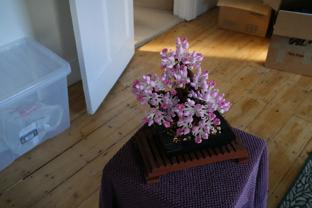}
     &  \includegraphics[width=1\linewidth]{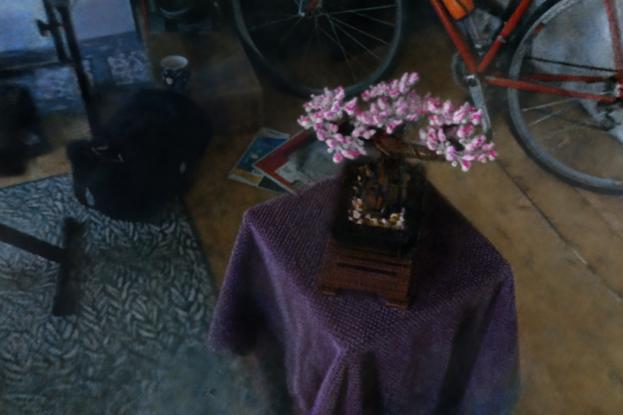}
     &  \includegraphics[width=1\linewidth]{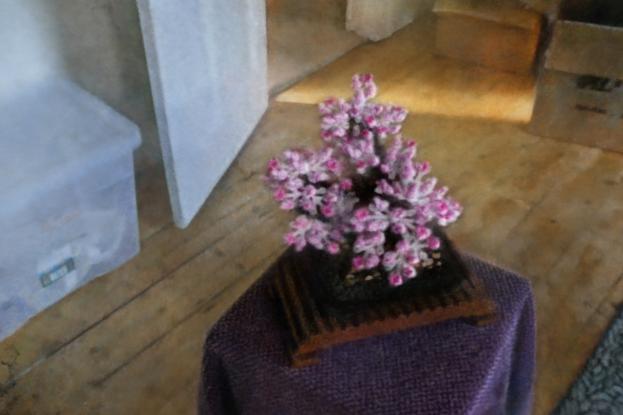}
     &  \includegraphics[width=1\linewidth]{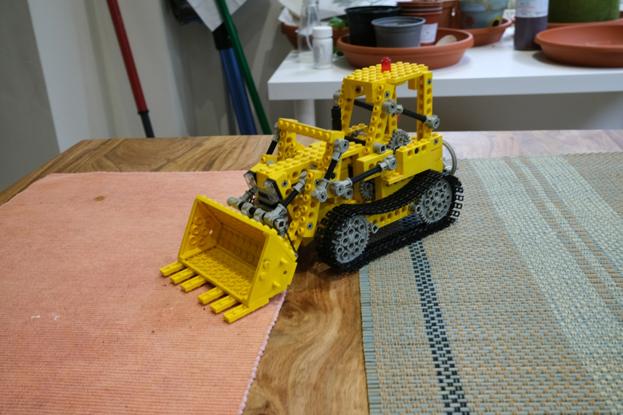}
     &  \includegraphics[width=1\linewidth]{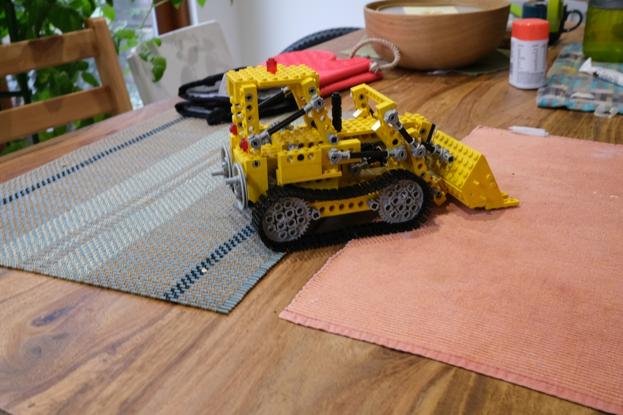}
     &  \includegraphics[width=1\linewidth]{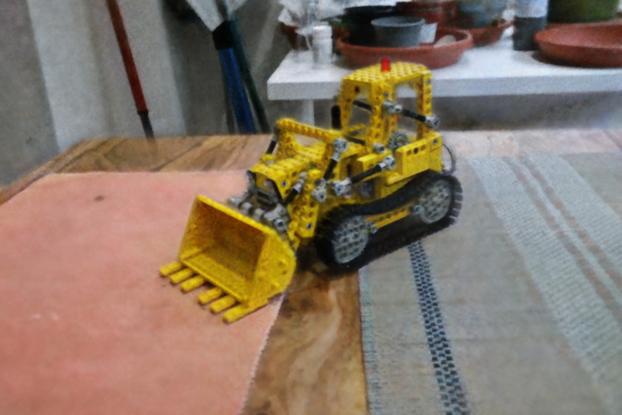}
     &  \includegraphics[width=1\linewidth]{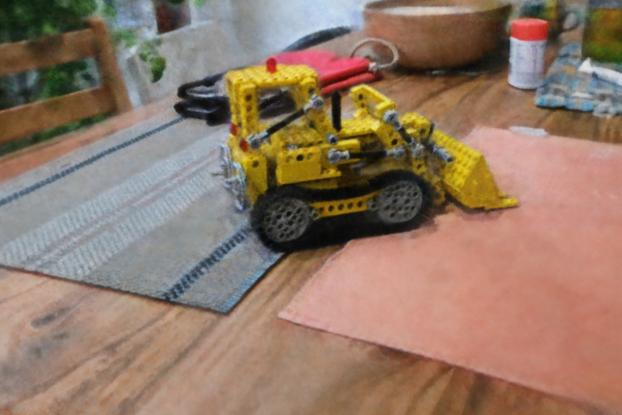} \\ 

     {Garden (Treehill) } 
     &  \includegraphics[width=1\linewidth]{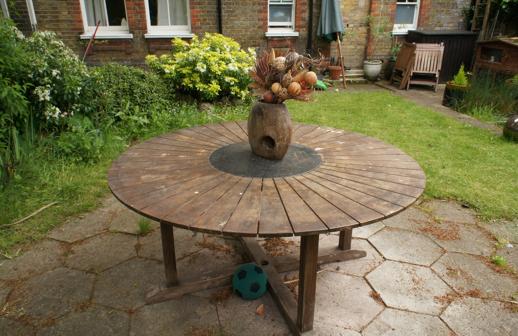}
     &  \includegraphics[width=1\linewidth]{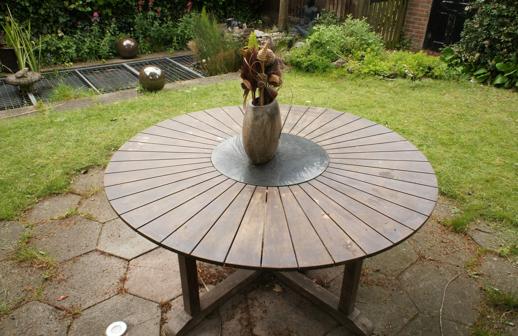}
     &  \includegraphics[width=1\linewidth]{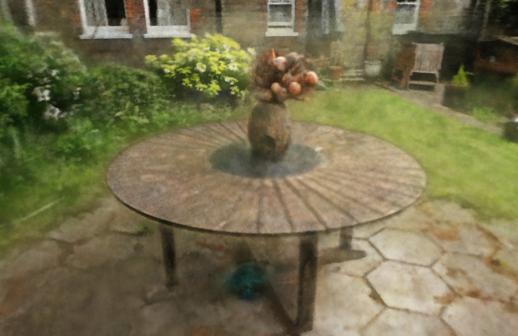}
     &  \includegraphics[width=1\linewidth]{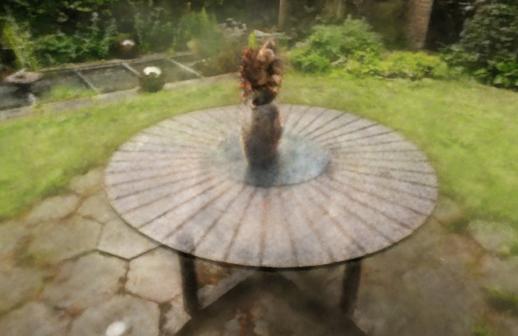}
     &  \includegraphics[width=1\linewidth]{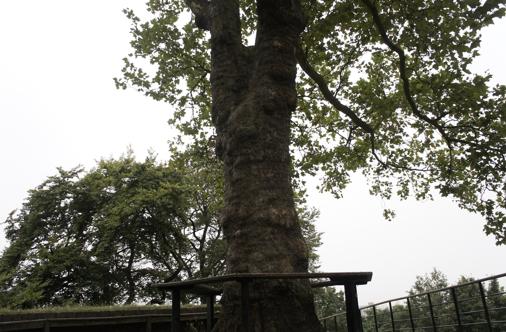}
     &  \includegraphics[width=1\linewidth]{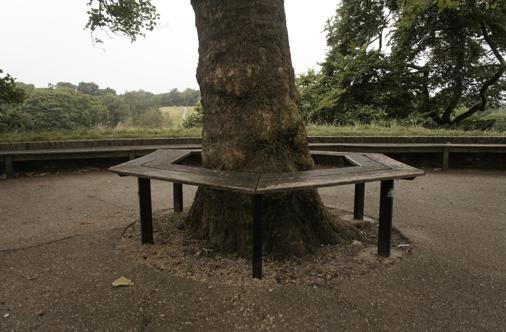}
     &  \includegraphics[width=1\linewidth]{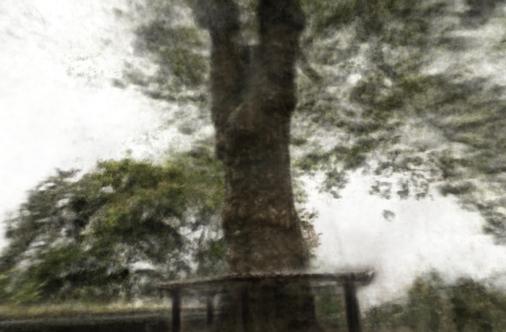}
     &  \includegraphics[width=1\linewidth]{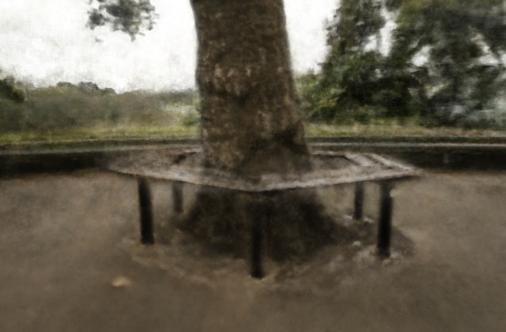} \\ 

     {Kitchen (Counter) } 
     &  \includegraphics[width=1\linewidth]{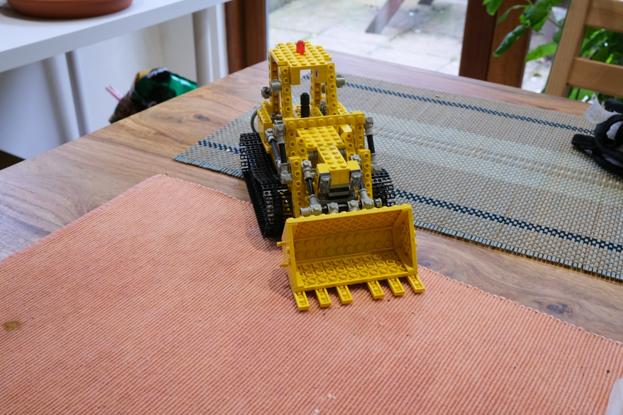}
     &  \includegraphics[width=1\linewidth]{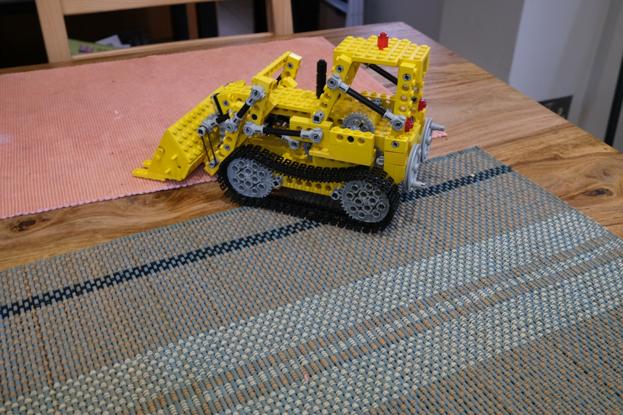}
     &  \includegraphics[width=1\linewidth]{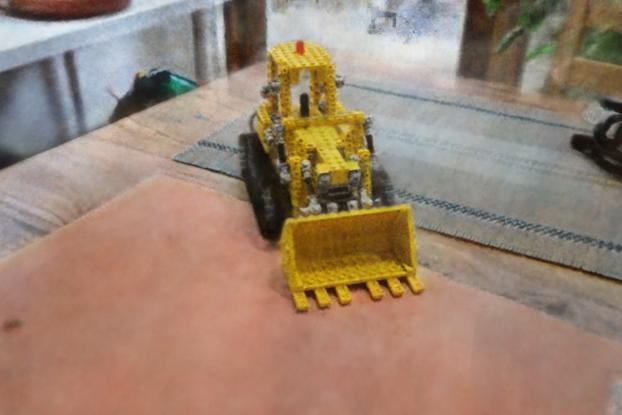}
     &  \includegraphics[width=1\linewidth]{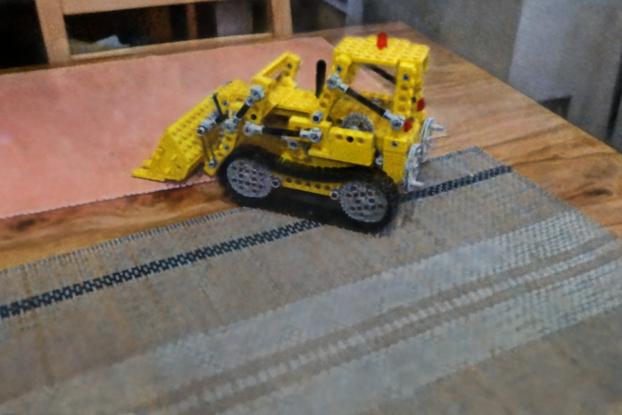}
     &  \includegraphics[width=1\linewidth]{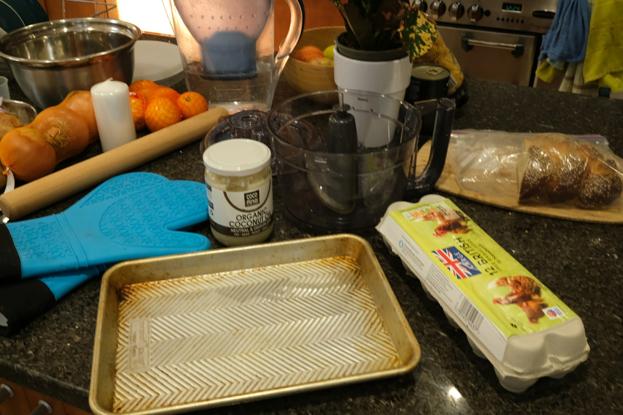}
     &  \includegraphics[width=1\linewidth]{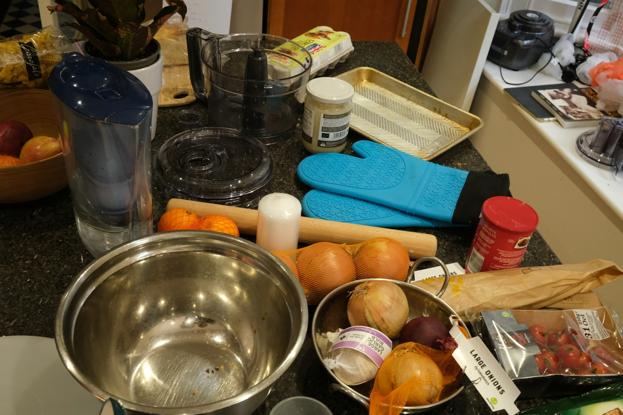}
     &  \includegraphics[width=1\linewidth]{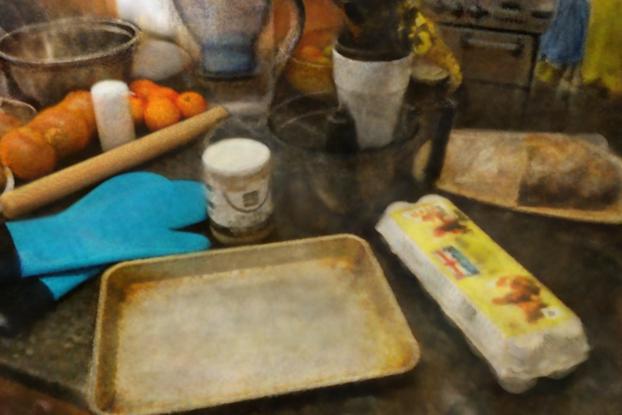}
     &  \includegraphics[width=1\linewidth]{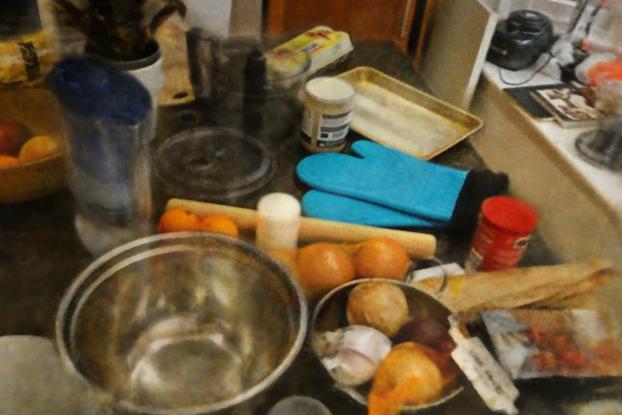} \\ 
     
     {Room (Bonsai)} 
     &  \includegraphics[width=1\linewidth]{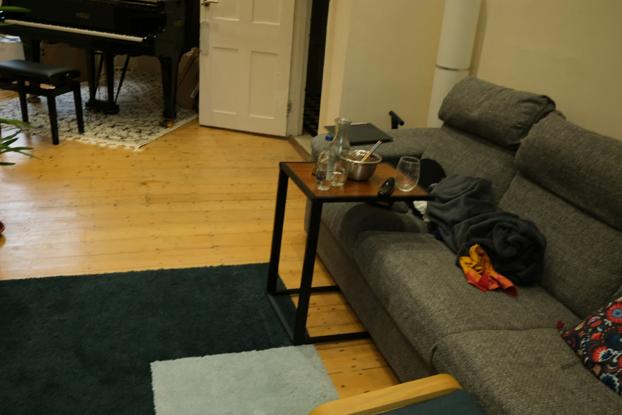}
     &  \includegraphics[width=1\linewidth]{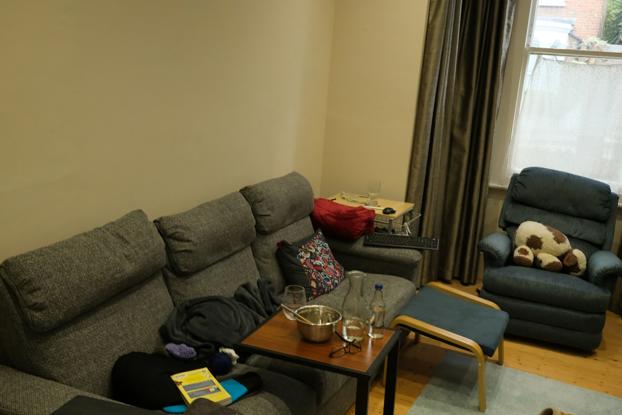}
     &  \includegraphics[width=1\linewidth]{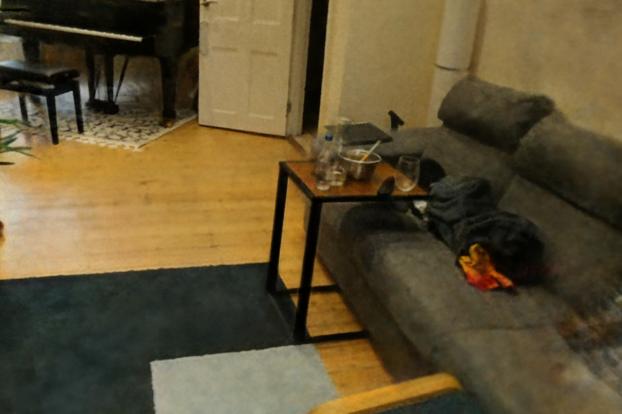}
     &  \includegraphics[width=1\linewidth]{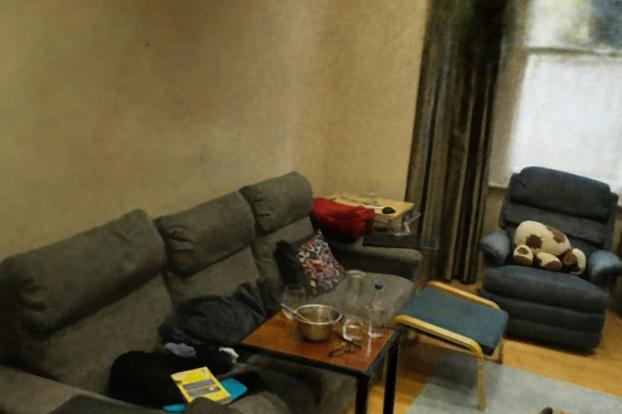}
     &  \includegraphics[width=1\linewidth]{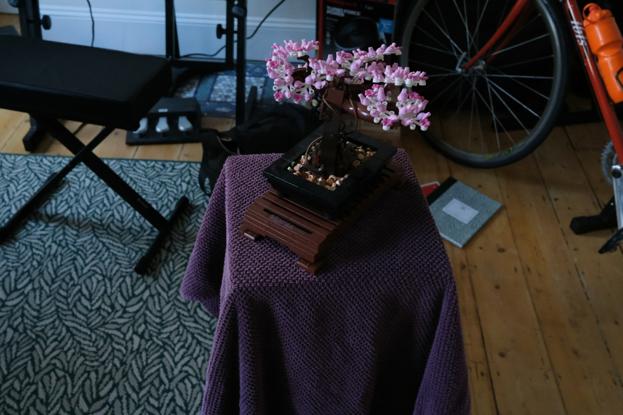}
     &  \includegraphics[width=1\linewidth]{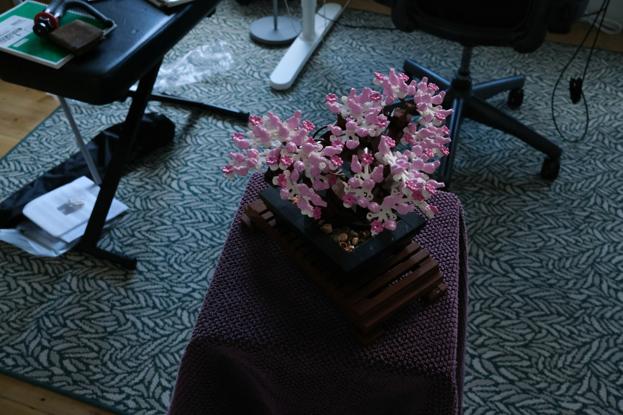}
     &  \includegraphics[width=1\linewidth]{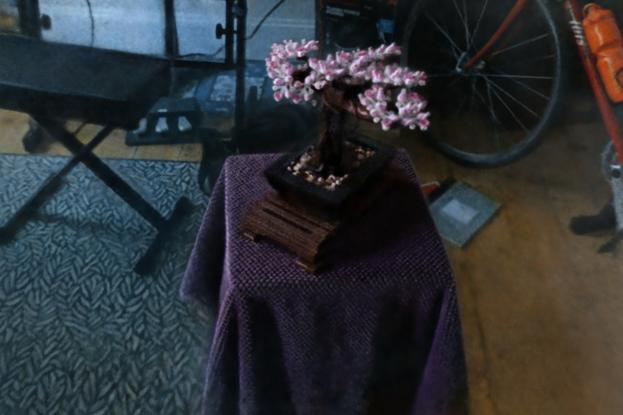}
     &  \includegraphics[width=1\linewidth]{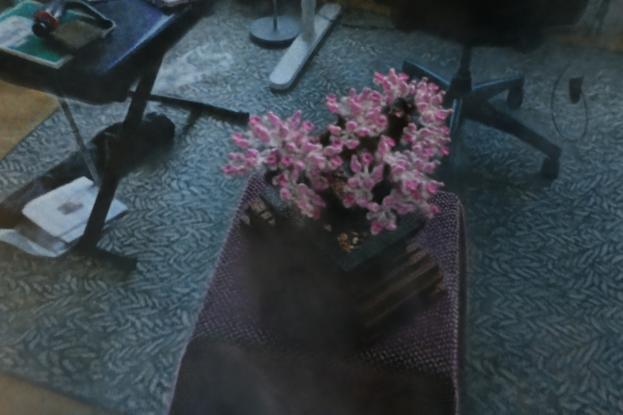} \\ 
     
    & \multicolumn{2}{c}{Ground Truth ($S$)} 
    & \multicolumn{2}{c}{\textbf{StegoNGP} ($\Pi$)}
    & \multicolumn{2}{c}{Ground Truth ($\mathcal{B}$)} 
    & \multicolumn{2}{c}{\textbf{StegoNGP} ($\mathcal{K}$)} \\
    
\end{tabular}
   \caption{Visualization of StegoNGP performance on the Mip-Nerf-360 dataset. The figure compares the rendered results for several scene pairs against their respective Ground Truth images. Pairs are denoted as Cover Scene (Hidden Scene).}
    \label{fig:visual_mipnerf360}
\end{figure*}

To evaluate the generalization capability of StegoNGP on complex, real-world data, we conducted experiments on the challenging Mip-Nerf-360 dataset. This dataset features unbounded, large-scale scenes with complex lighting. We provide the visualization of our StegoNGP performance 
\wjx{as illustrated} 
in Figure \ref{fig:visual_mipnerf360}, comparing the rendered results against the Ground Truth.

As shown in Figure \ref{fig:visual_mipnerf360}, StegoNGP successfully learns to represent both the complex cover scene and the embedded hidden scene. This demonstrates the feasibility of our high-capacity, cryptographic steganography method beyond synthetic datasets and its applicability to real-world scenarios.

\subsection{Ablation Study on the Multi-Key Scheme} \label{sec:sec_Multi_Key}


\begin{figure}[htp]
\scriptsize
\centering
\setlength{\tabcolsep}{1pt}
\begin{tabular}{m{1.9cm}m{2.0cm}m{2.0cm}m{2.0cm}}

     &  \multicolumn{1}{c}{$m=4$}
     &  \multicolumn{1}{c}{$m=8$}
     &  \multicolumn{1}{c}{$m=16$} \\

        $0\%$ Secret Keys ($100\%$ Default)
     &  \includegraphics[width=1\linewidth]{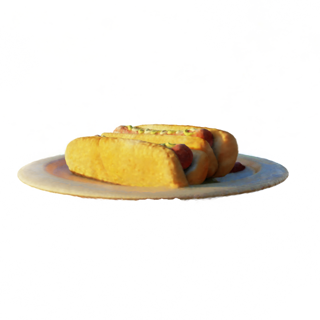}
     &  \includegraphics[width=1\linewidth]{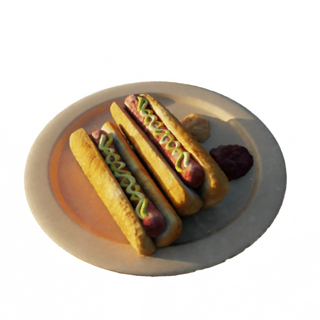}
     &  \includegraphics[width=1\linewidth]{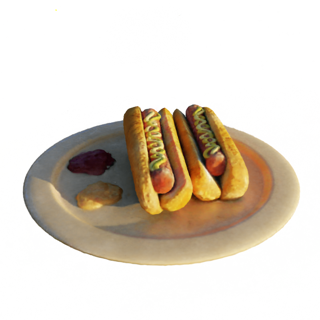}\\

        $25\%$ Secret Keys ($75\%$ Default)
     &  \includegraphics[width=1\linewidth]{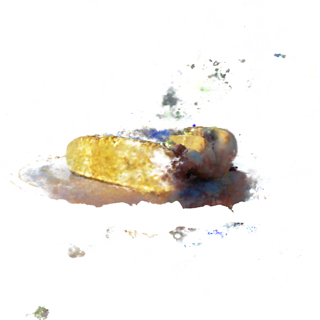}
     &  \includegraphics[width=1\linewidth]{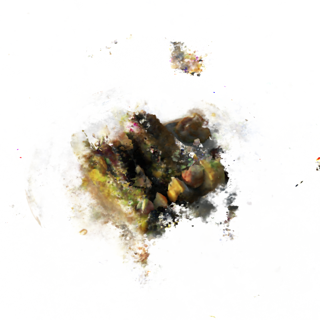}
     &  \includegraphics[width=1\linewidth]{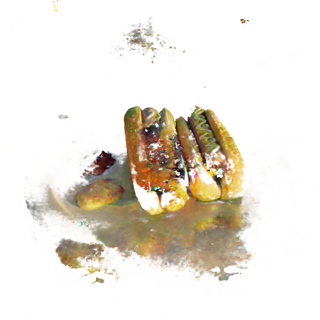}\\

        $50\%$ Secret Keys ($50\%$ Default)
     &  \includegraphics[width=1\linewidth]{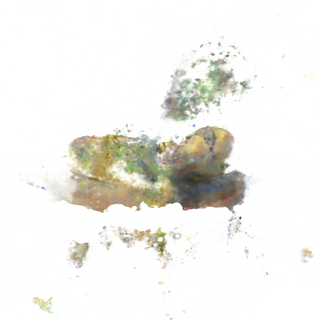}
     &  \includegraphics[width=1\linewidth]{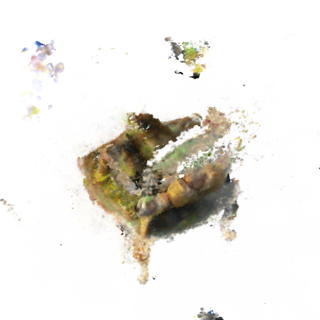}
     &  \includegraphics[width=1\linewidth]{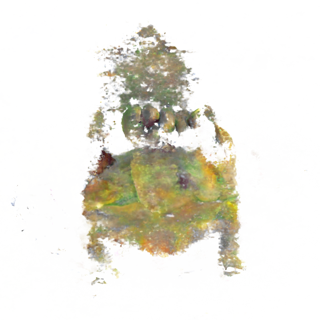}\\

        $75\%$ Secret Keys ($25\%$ Default)
     &  \includegraphics[width=1\linewidth]{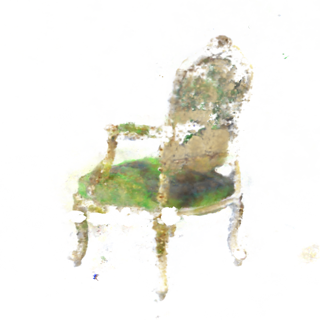}
     &  \includegraphics[width=1\linewidth]{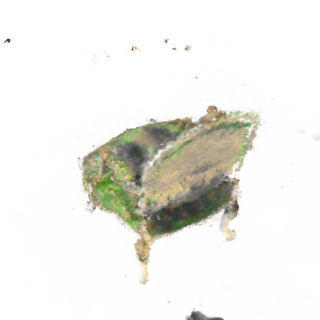}
     &  \includegraphics[width=1\linewidth]{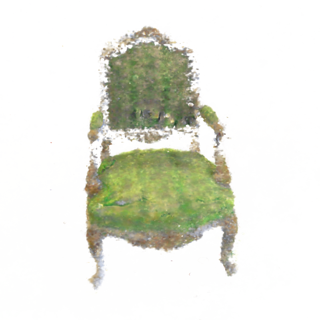}\\

        $100\%$ Secret Keys ($0\%$ Default)
     &  \includegraphics[width=1\linewidth]{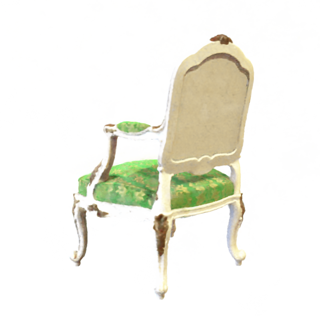}
     &  \includegraphics[width=1\linewidth]{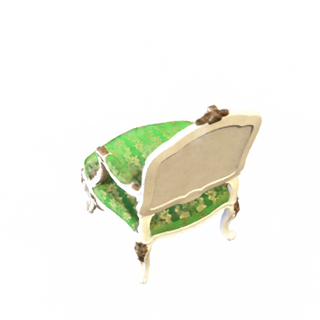}
     &  \includegraphics[width=1\linewidth]{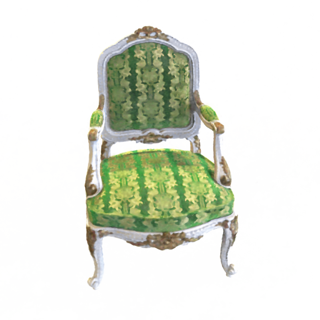}\\


        \hspace{0.3cm}$\textbf{K}$ Provision
     &  \multicolumn{3}{c}{rendered image (test set)}\\
     
\end{tabular}
   \caption{The Multi-Key StegoNGP scheme and its robustness against partial key disclosure. Columns represent different total key quantities ($m=4, 8, 16$). Rows show the rendered results as the proportion of the correct secret key set $\mathbf{K}$ increases from $0\%$ to $100\%$ (where missing secret keys are substituted with the default key $\Pi$). The model transitions from the cover scene $\mathcal{S}$ ($0\%$ $\textbf{K}$ provision) to the hidden scene $\mathcal{B}$ ($100\%$ $\textbf{K}$ provision).}
    \label{fig:visual_multi_key}
\end{figure}

In this subsection, we analyze the effectiveness of the Multi-Key StegoNGP scheme introduced in Section \ref{subsec:multi_key}. A key motivation for this scheme (where $m>1$) is to significantly enhance the cryptographic security compared to the standard ($m=1$) configuration. We investigate the robustness of this multi-key approach by simulating a scenario where an external observer possesses an incomplete set of secret keys. In this experiment, a varying proportion of the correct secret key set $\mathbf{K}$ is supplied to the model, while the remaining key slots are 
\jwx{filled with} the default public key $\Pi$. This 
\wjx{evaluates}
the Multi-Key scheme resilience and security threshold against partial key disclosure attacks.

\wjx{From }
Figure \ref{fig:visual_multi_key}, 
\wjx{we can see that the results}
demonstrate a high security threshold \wjx{of the Multi-Key scheme}. 
At $0\%$ secret key provision, the model correctly defaults to rendering the cover scene $\mathcal{S}$, which establishes plausible deniability. Even with significant partial key knowledge ($e.g.$, $50\%$), the model fails to reconstruct the hidden scene, instead producing a 
\jwx{incoherent output.} While some features of the hidden scene may become faintly perceptible at high provision levels ($e.g.$, $75\%$), a coherent and usable reconstruction is only achieved when the entire $100\%$ of the secret key set is supplied. This analysis demonstrates that the Multi-Key scheme effectively protects the hidden scene $\mathcal{B}$, as an attacker must possess the complete set of $m$ keys for a successful recovery.


\section{Conclusion}

In this paper, \wjx{we introduce a novel 3D cryptographic steganography method using Instant-NGP (StegoNGP) for addressing} the challenge of high-capacity data hiding with zero parameter overhead. Our core contribution is the transformation of the Instant-NGP hash function into a key-controlled scene switcher, allowing a single model to embed both a public cover scene and a secret hidden scene using distinct cryptographic keys. Experiments \wjx{results} demonstrated that our method achieves excellent imperceptibility, rendering a cover scene indistinguishable from a standard baseline, whereas other methods remain structurally detectable. Furthermore, we introduced an enhanced Multi-Key StegoNGP scheme that dramatically expands the key space. Ablation studies confirmed this scheme is highly resistant to partial key disclosure, ensuring the hidden scene is accessible with the complete and correct key set.

\newpage

{
    \small
    \bibliographystyle{ieeenat_fullname}
    \bibliography{main}
}

\clearpage
\setcounter{page}{1}
\maketitlesupplementary

This supplementary document provides additional details, quantitative analysis, and ablation studies to support the claims made in the main paper. We first present a detailed Security Analysis, offering a quantitative breakdown of the cryptographic key space and robustness against brute-force attacks. We then provide an Analysis of Hash Collisions and Table Size, which investigates the impact of hash table capacity on reconstruction fidelity in complex scenes. Following this, we present the complete, per-scene Detailed Experimental Results for all experiments summarized in the main paper, including the synthetic dataset, real-world dataset, comparison with existing methods, and the detailed metrics for the Multi-Key ablation study. Finally, we include an additional experiment on Embedding Multiple Hidden Scenes to explore the capacity limits of our method.

\setcounter{figure}{6} 
\setcounter{table}{2}     

\section{Security Analysis}
This section provides a quantitative analysis of the StegoNGP scheme's key space and its robustness against brute-force attacks. We assume an attacker has 
\jwx{full model and algorithmic knowledge}, but not the secret keys ($\mathcal{K}$ or $\mathbf{K}$), and must 
\jwx{exhaustively search all key combinations.}

\subsection{Key Space Analysis}
Our \textbf{Basic StegoNGP} method ($m=1$) uses 3-prime tuples ($\mathcal{K}\equiv\{k_1, k_2, k_3\}$) selected from a prime pool in range $[10^7, 10^{10}]$. Let $P(x)$ be the prime-counting function for the number of primes in the range $(0, x)$, the total number of available primes $U$ in this range is:

\begin{align}
\begin{split}
    U & = P(10^{10}) - P(10^7) \\
    & = 455,052,511 - 664,579 = 454,387,932,
\end{split}
\end{align}
the total key space $Q_{basic}$ and its equivalent bit strength are:
\begin{align}
    Q_{basic} = U^3 & = (454,387,932)^3 \approx \num{9.364e25}, \\
    \text{Bit Strength} & = \log_2(Q_{basic}) \approx \mathbf{86.3 \text{ bits}},
\end{align}
a $2^{86}$ key space is computationally secure. For the \textbf{Multi-Key StegoNGP} ($m=16$), an attacker must find all $16 \times 3$ prime keys (48 primes total). The total key space $Q_{multi}$ is:
\begin{align}
    Q_{multi} &= (Q_{basic})^{16} = (U^3)^{16} \nonumber, \\
    \text{Bit Strength} &= \log_2(Q_{multi}) \approx 16 \times (86.3) \approx \mathbf{1380 \text{ bits}},
\end{align}
a $2^{1380}$ key space is astronomically larger than cryptographic standards like AES-256 and is computationally infeasible to break with current computing power.

\subsection{Brute-Force Feasibility}

We analyze the feasibility of breaking the \textbf{basic model} ($Q_{basic} \approx \num{9.36e25}$) under assumptions highly favorable to the attacker:
(1) high-performance GPU with sufficient VRAM, such as a NVIDIA RTX 5090 (32GB of VRAM and 3,352 TOPS of computing power), (2) the render speed of 20 images per second (20 FPS), and (3) 
\jwx{an ideal}, instantaneous classifier to distinguish coherent scenes from noise.

The average time to crack ($T_{\text{years}}$), assuming a 50\% search of the key space, is:
\begin{align}
    T_{\text{sec}} &= \frac{Q_{basic} \times 0.5}{20} \approx \num{2.341e24} \text{ seconds} \\
    T_{\text{years}} &= \frac{\num{2.341e24} \text{ sec}}{3.15 \times 10^7 \text{ sec/year}} \approx \num{7.43e16} \text{ years}
\end{align}
Even under these 
\jwx{ideal} assumptions, breaking our basic model would require approximately \textbf{74.3 quadrillion years}, which is 
\jwx{computationally infeasible}. Our quantitative analysis, summarized in Table \ref{tab:security_summary}, confirms the cryptographic robustness of the StegoNGP scheme.

\begin{table*}[htbp!]
\centering
\caption{StegoNGP Key Space and Security Summary}
\label{tab:security_summary}
\resizebox{\textwidth}{!}{%
\begin{tabular}{@{}l l l l l@{}}
\toprule
Model & Key Composition & Key Space Size & Bit Strength & Est. Brute-Force Time (RTX 5090 @ 20 FPS) \\ \midrule
\textbf{Basic} ($m=1$) & 3 primes & $\approx \num{9.36e25}$ & $\approx 2^{86.3}$ & $\approx \num{7.43e16}$ years \\
\textbf{Multi-Key} ($m=16$) & 48 primes & $\approx \num{3.46e415}$ & $\approx 2^{1380}$ & Computationally Infeasible \\ 
AES-256 \cite{wright2001advanced} & 256-bit key & $\approx \num{1.15e77}$ & $2^{256}$ & Computationally Infeasible \\ \bottomrule
\end{tabular}%
}
\end{table*}

\section{Analysis of Hash Collisions and Table Size} \label{sec:sec_Hash_Collisions}
In our experiments, we observed a performance gap between the simple synthetic datasets and the complex, large-scale real-world datasets like Mip-Nerf-360 (detailed results are shown in Table \ref{tab:supp_real_detailed}). We hypothesize that this degradation is primarily due to an increased rate of hash collisions within the Instant-NGP multi-resolution hash encoding's feature table.

\subsection{Collision Analysis}
Instant-NGP achieves its efficiency by mapping coordinates from sparse, high-resolution grids to a dense feature table $T$ of a fixed size. As noted in the Instant-NGP paper, 3D space is sparse; for example, a grid with resolution $128^3$ (over 2 million cells) might only have about $54,000$ active cells that actually contribute to the visible surfaces.

In the standard Instant-NGP, these $\kappa_1 \approx 54,000$ active cells are mapped via a hash function into the feature table $T$. Our model uses the default table size $T = 2^{19} = 524,288$. This creates, ideally, a fill rate of $\rho_1 = \kappa_1 / T \approx 54,000 / 524,288 \approx 10.3\%$. While some collisions are inevitable, the sparsity and the large table size keep them at a manageable level for a single scene. Therefore, the original Instant-NGP paper \cite{muller2022instant} did not explicitly handle hash collisions.

However, our StegoNGP method interweaves two complete 3D scenes (Cover $S$ and Hidden $\mathcal{B}$) into the \textit{same} feature table $T$. The cover scene $S$, using the default key $\Pi$, maps its $\kappa_1$ active cells to $T$. Simultaneously, the hidden scene $\mathcal{B}$, using the secret key $\mathcal{K}$, maps its $\kappa_2$ active cells to the same $T$.

Assuming the hidden scene has similar complexity (e.g., $\kappa_2 \approx 54,000$), the total number of active cells being mapped to $T$ is now $\kappa_{total} = \kappa_1 + \kappa_2 \approx 108,000$. This effectively doubles the table's fill rate to $\rho_{stego} = \kappa_{total} / T \approx 20.6\%$. This significantly increases the probability of hash collisions, where feature vectors from scene $S$ and scene $\mathcal{B}$ are hashed to the same table entry, creating feature interference during training. This interference forces the network to find a suboptimal compromise for the feature vector at that slot, degrading the reconstruction quality for both scenes.

\subsection{Experimental Verification}
To test this hypothesis, we designed an experiment to measure the effect of hash table size $T$ on reconstruction fidelity for a complex, real-world scene pair. We used the "Bonsai" (Cover $S$) and "Kitchen" (Hidden $\mathcal{B}$) pair from the Mip-Nerf-360 dataset and trained our StegoNGP model with three different hash table sizes:
\begin{itemize}
    \item $T = 2^{18}$ (Half: 262,144 entries);
    \item $T = 2^{19}$ (Default: 524,288 entries);
    \item $T = 2^{20}$ (Double: 1,048,576 entries);
    \item $T = 2^{21}$ (Quadruple: 2,097,152 entries).
\end{itemize}

The results are presented in Table \ref{tab:supp_ablation_hash_size}.

\begin{table}[htbp!]
    \centering
    \caption{Quantitative results of varying the hash table size $T$ on the Mip-Nerf-360 "Bonsai" ($S$) / "Kitchen" ($\mathcal{B}$) scene pair. As $T$ increases, the reduced hash collision probability leads to a clear improvement in reconstruction fidelity for both the cover and hidden scenes.}
    \label{tab:supp_ablation_hash_size}
    \begin{tabular}{@{}c l c c c@{}}
        \toprule
        Table Size ($T$) & Scene & PSNR $\uparrow$ & SSIM $\uparrow$ & LPIPS $\downarrow$ \\
        \midrule
        \multirow{2}{*}{$T = 2^{18}$} & $S$ & 23.145 & 0.715 & 0.317 \\
         & $\mathcal{B}$ & 22.797 & 0.572 & 0.393 \\
        \midrule
        \multirow{2}{*}{\textbf{$T = 2^{19}$}} & $S$ & 22.862 & 0.721 & 0.286 \\
         & $\mathcal{B}$ & 22.888 & 0.575 & 0.398 \\
        \midrule
        \multirow{2}{*}{$T = 2^{20}$} & $S$ & 23.626 & 0.754 & 0.235 \\
         & $\mathcal{B}$ & 22.695 & 0.579 & 0.388 \\
        \midrule
        \multirow{2}{*}{$T = 2^{21}$} & $S$ & 22.998 & 0.753 & 0.228 \\
         & $\mathcal{B}$ & 22.941 & 0.621 & 0.343 \\
        \bottomrule
    \end{tabular}
\end{table}

The results in Table \ref{tab:supp_ablation_hash_size} strongly support our hypothesis. By doubling ($T=2^{20}$) and quadrupling ($T=2^{21}$) the hash table size, we provide more space for the two scenes' feature vectors, significantly reducing the rate of collision and feature interference. This leads to a consistent and noticeable improvement in PSNR, SSIM, and LPIPS for both the cover and hidden scenes. Conversely, when we halve the hash table ($T=2^{18}$), the reconstruction quality decreases.   
\section{Detailed Experimental Results}
This section provides a detailed breakdown of the quantitative metrics for the experiments presented in the main paper, including the full results for the synthetic dataset (Sec \ref{sec:sec_Synthetic_Scene}), the per-pair comparison against existing methods (Sec \ref{sec:sec_Existing_Methods}), the quantitative metrics for the real-world dataset (Sec \ref{sec:sec_Real_World}), and the detailed data for the multi-key ablation study (Sec \ref{sec:sec_Multi_Key}).

\subsection{Detailed Performance on Synthetic Scene (Sec \ref{sec:sec_Synthetic_Scene})}
Table \ref{tab:supp_synthetic_detailed} expands on Table \ref{table:main_experiment} from the main paper. It shows the individual reconstruction fidelity (PSNR, SSIM, LPIPS) for all eight scene pairs used in the Blender Synthetic dataset experiment. The \textit{Average} row corresponds to the \textit{StegoNGP (Ours)} row in the main paper's Table \ref{table:main_experiment}.

\begin{table*}[htbp!]
    \centering
    \caption{Detailed quantitative results on the Blender Synthetic dataset. Each row represents a StegoNGP model trained to hide one scene within another. The metrics show the fidelity for rendering the cover scene $S$ (using key $\Pi$) and the hidden scene $\mathcal{B}$ (using key $\mathcal{K}$).}
    \label{tab:supp_synthetic_detailed}
    \begin{tabular}{@{}l l c c c c c c@{}}
        \toprule
        \multicolumn{2}{c}{Scene Pair} & \multicolumn{3}{c}{Cover Scene ($S$)} & \multicolumn{3}{c}{Hidden Scene ($\mathcal{B}$)} \\
        Cover Scene ($S$) & Hidden Scene ($\mathcal{B}$) & PSNR $\uparrow$ & SSIM $\uparrow$ & LPIPS $\downarrow$ & PSNR $\uparrow$ & SSIM $\uparrow$ & LPIPS $\downarrow$ \\
        \midrule
        Lego & Twitter & 26.902 & 0.910 & 0.070 & 32.514 & 0.985 & 0.009 \\
        Mic & Twitter & 28.361 & 0.955 & 0.052 & 32.693 & 0.984 & 0.010 \\
        Chair & Twitter & 28.308 & 0.921 & 0.086 & 32.630 & 0.986 & 0.009 \\
        Ficus & Twitter & 26.950 & 0.935 & 0.044 & 32.891 & 0.985 & 0.009 \\
        Hotdog & Google & 29.586 & 0.946 & 0.056 & 34.991 & 0.991 & 0.006 \\
        Drums & Google & 22.730 & 0.894 & 0.103 & 35.804 & 0.992 & 0.005 \\
        Ship & Google & 24.847 & 0.814 & 0.181 & 34.902 & 0.991 & 0.006 \\
        Materials & Google & 24.965 & 0.903 & 0.070 & 35.957 & 0.992 & 0.005 \\
        \textbf{Average} & & \textbf{26.581} & \textbf{0.910} & \textbf{0.083} & \textbf{34.048} & \textbf{0.988} & \textbf{0.007} \\
        \bottomrule
    \end{tabular}
    \vspace{0.2cm}
\end{table*}

\begin{table*}[htbp!]
    \centering
    \caption{Detailed results for the \textbf{GS-Hider \cite{zhang2024gs}} method across the four specialized scene pairs. The average of these metrics corresponds to the GS-Hider row in Table \ref{table:compared_with_gs_hider} of the main paper.}
    \label{tab:supp_comparison_gshider}
    \begin{tabular}{@{}l l c c c c c c@{}}
        \toprule
        \multicolumn{2}{c}{Scene Pair} & \multicolumn{3}{c}{Cover Scene ($S$)} & \multicolumn{3}{c}{Hidden Scene ($\mathcal{B}$)} \\
        Cover Scene ($S$) & Hidden Scene ($\mathcal{B}$) & PSNR $\uparrow$ & SSIM $\uparrow$ & LPIPS $\downarrow$ & PSNR $\uparrow$ & SSIM $\uparrow$ & LPIPS $\downarrow$ \\
        \midrule
        Hotdog & Lego & 35.376 & 0.985 & 0.009 & 30.048 & 0.971 & 0.016 \\
        Mic & Drums & 36.601 & 0.993 & 0.005 & 27.091 & 0.959 & 0.028 \\
        Chair & Ship & 37.055 & 0.992 & 0.007 & 29.460 & 0.931 & 0.027 \\
        Ficus & Materials & 35.744 & 0.995 & 0.004 & 31.899 & 0.985 & 0.049 \\
        \textbf{Average} & & \textbf{36.194} & \textbf{0.991} & \textbf{0.006} & \textbf{29.625} & \textbf{0.962} & \textbf{0.030} \\
        \bottomrule
    \end{tabular}
    \vspace{0.2cm}
\end{table*}

\begin{table*}[htbp!]
    \centering
    \caption{Detailed quantitative results for our \textbf{StegoNGP (Ours)} method across the four specialized scene pairs. The average of these metrics corresponds to the StegoNGP (Ours) row in Table \ref{table:compared_with_gs_hider} of the main paper.}
    \label{tab:supp_comparison_stegongp}
    \begin{tabular}{@{}l l c c c c c c@{}}
        \toprule
        \multicolumn{2}{c}{Scene Pair} & \multicolumn{3}{c}{Cover Scene ($S$)} & \multicolumn{3}{c}{Hidden Scene ($\mathcal{B}$)} \\
        Cover Scene ($S$) & Hidden Scene ($\mathcal{B}$) & PSNR $\uparrow$ & SSIM $\uparrow$ & LPIPS $\downarrow$ & PSNR $\uparrow$ & SSIM $\uparrow$ & LPIPS $\downarrow$ \\
        \midrule
        Hotdog & Lego & 32.269 & 0.953 & 0.052 & 28.100 & 0.923 & 0.062 \\
        Mic & Drums & 29.213 & 0.957 & 0.050 & 24.017 & 0.907 & 0.093 \\
        Chair & Ship & 28.796 & 0.925 & 0.085 & 26.184 & 0.832 & 0.166 \\
        Ficus & Materials & 28.434 & 0.948 & 0.036 & 26.689 & 0.926 & 0.056 \\
        \textbf{Average} & & \textbf{29.678} & \textbf{0.946} & \textbf{0.056} & \textbf{26.248} & \textbf{0.897} & \textbf{0.094} \\
        \bottomrule
    \end{tabular}
    \vspace{0.2cm}
\end{table*}

\begin{table*}[htbp!]
    \centering
    \caption{Detailed results for StegoNGP on the Mip-Nerf-360 dataset. Each row corresponds to a scene pair visualized in Figure \ref{fig:visual_mipnerf360}.}
    \label{tab:supp_real_detailed}
    \begin{tabular}{@{}l l c c c c c c@{}}
        \toprule
        \multicolumn{2}{c}{Scene Pair} & \multicolumn{3}{c}{Cover Scene ($S$)} & \multicolumn{3}{c}{Hidden Scene ($\mathcal{B}$)} \\
        Cover Scene ($S$) & Hidden Scene ($\mathcal{B}$) & PSNR $\uparrow$ & SSIM $\uparrow$ & LPIPS $\downarrow$ & PSNR $\uparrow$ & SSIM $\uparrow$ & LPIPS $\downarrow$ \\
        \midrule
        Bonsai & Kitchen & 22.862 & 0.721 & 0.286 & 22.888 & 0.575 & 0.398 \\
        Garden & Treehill & 20.559 & 0.405 & 0.578 & 16.951 & 0.430 & 0.545 \\
        Kitchen & Counter & 21.289 & 0.550 & 0.425 & 21.854 & 0.678 & 0.397 \\
        Room & Bonsai & 24.696 & 0.789 & 0.259 & 21.995 & 0.707 & 0.264 \\
        \textbf{Average} & & \textbf{22.352} & \textbf{0.616} & \textbf{0.387} & \textbf{20.922} & \textbf{0.598} & \textbf{0.401} \\
        \bottomrule
    \end{tabular}
    \vspace{0.2cm}
\end{table*}

\subsection{Detailed Comparison with Existing Methods (Sec \ref{sec:sec_Existing_Methods})}
Tables \ref{tab:supp_comparison_gshider} and \ref{tab:supp_comparison_stegongp} provide the per-pair breakdown for the comparison summarized in Table \ref{table:compared_with_gs_hider} of the main paper. This experiment used four scene pairs with aligned camera poses, as described in Sec \ref{sec:sec_Existing_Methods}. Table \ref{tab:supp_comparison_gshider} shows the results for GS-Hider \cite{zhang2024gs}, and Table \ref{tab:supp_comparison_stegongp} shows the results for our StegoNGP.

\subsection{Detailed Performance on Real World Scene (Sec \ref{sec:sec_Real_World})}
Table \ref{tab:supp_real_detailed} provides the quantitative metrics (PSNR, SSIM, LPIPS) for the experiments on the Mip-Nerf-360 dataset \cite{barron2022mip}. These metrics correspond to the qualitative visualizations presented in Figure \ref{fig:visual_mipnerf360} of the main paper.

\subsection{Detailed Ablation Study on the Multi-Key Scheme (Sec \ref{sec:sec_Multi_Key})}
This section provides the detailed data corresponding to the ablation study in Section \ref{sec:sec_Multi_Key} of the main paper. We first list the set of 16 secret keys (Table \ref{tab:supp_key_list}) used for the multi-key experiments, from which the first 4 keys are used for $m=4$, the first 8 keys for $m=8$, and all 16 keys for $m=16$. We then provide the quantitative metrics (Table \ref{tab:supp_ablation_partial}) that support the partial key disclosure analysis in Figure \ref{fig:visual_multi_key}, showing the reconstruction fidelity as the proportion of correct secret keys increases.

\begin{table}[htbp!]
    \centering
    \caption{The master set of 16 secret keys used in the Multi-Key StegoNGP experiments. Each key $\mathcal{K}_j$ is a 3-prime tuple $\{k_{j,1}, k_{j,2}, k_{j,3}\}$.}
    \label{tab:supp_key_list}
    \begin{tabular}{@{}c c c c@{}}
        \toprule
        Key ID & Prime & Prime & Prime \\
        ($j$) & $k_{j,1}$ & $k_{j,2}$ & $k_{j,3}$ \\
        \midrule
            1 & 10123105283 & 9423208789 & 10600480711 \\
            2 & 3674653429 & 2097192037 & 1434869437 \\
            3 & 8542170799 & 6143207549 & 6519670961 \\
            4 & 15443898139 & 12668029511 & 12301111429 \\
            5 & 14405672371 & 5760082793 & 11538455387 \\
            6 & 13705036483 & 5104709759 & 3761188213 \\
            7 & 3364878689 & 5035457381 & 1560686051 \\
            8 & 10808744237 & 9293063341 & 12211915819 \\
            9 & 9320630941 & 2357489971 & 12349316797 \\
            10 & 5123446871 & 10280445341 & 8301841781 \\
            11 & 5489304827 & 13394370001 & 11743146431 \\
            12 & 13560374447 & 8074960889 & 6193177019 \\
            13 & 3510659183 & 2993894131 & 8230892477 \\
            14 & 5512642787 & 4108776493 & 15236738411 \\
            15 & 8923388921 & 1689310319 & 8101897283 \\
            16 & 8288642353 & 3761544863 & 7037980441 \\
        \bottomrule
    \end{tabular}
\end{table}

\begin{table}[htbp!]
    \centering
    \caption{Detailed results for the Multi-Key StegoNGP partial key disclosure attack, corresponding to Figure \ref{fig:visual_multi_key} in the main paper (using Hotdog as Cover Scene $S$, Chair as Hidden Scene $\mathcal{B}$). We measure the reconstruction fidelity when the percentage of correct secret keys ($\mathbf{K}$) increases.}
    \label{tab:supp_ablation_partial}
    \begin{tabular}{@{}c c c c@{}}
        \toprule
        Model & K Provision (\%) & PSNR ($S$) & PSNR ($\mathcal{B}$) \\
        \midrule
        \multirow{5}{*}{\textbf{$m=4$}} & 0\% (Renders $S$) & \textbf{30.620} & 11.805\\
         & 25\% & 15.860 & 13.761 \\
         & 50\% & 13.822 & 16.219 \\
         & 75\% & 12.640 & 19.613 \\
         & 100\% (Renders $\mathcal{B}$) & 11.876 & \textbf{27.332} \\
        \midrule
        \multirow{5}{*}{\textbf{$m=8$}} & 0\% (Renders $S$) & \textbf{30.093} & 11.680 \\
         & 25\% & 13.421 & 14.255 \\
         & 50\% & 12.909 & 15.746 \\
         & 75\% & 12.243 & 19.548 \\
         & 100\% (Renders $\mathcal{B}$) & 11.857 & \textbf{27.513} \\
        \midrule
        \multirow{5}{*}{\textbf{$m=16$}} & 0\% (Renders $S$) & \textbf{30.556} & 11.850 \\
         & 25\% & 14.551 & 14.775 \\
         & 50\% & 12.676 & 18.193 \\
         & 75\% & 12.144 & 21.404\\
         & 100\% (Renders $\mathcal{B}$) & 11.877 & \textbf{27.731} \\
        \bottomrule
    \end{tabular}
\end{table}   
\section{Embedding Multiple Hidden Scenes}

To further explore the upper capacity limit of StegoNGP, we designed an experiment to embed multiple hidden scenes into a single cover scene model. The main paper demonstrates the $\Phi=1$ case, hiding one complete 3D scene. Here, we extend this to $\Phi=2, 3,$ and $4$ hidden scenes.

This is achieved by extending the training logic from Algorithm 1. Instead of randomly choosing between $S$ (using key $\Pi$) and $\mathcal{B}$ (using key $\mathcal{K}$), we randomly sample from the set $\{S, \mathcal{B}_1, \mathcal{B}_2, ..., \mathcal{B}_\Phi\}$, where each scene $\mathcal{B}_i$ is assigned a unique secret key $\mathcal{K}_i$.

We hypothesize that as $\Phi$ increases, the reconstruction quality for all scenes will degrade. This is a direct consequence of the hash collision problem analyzed in the section~\ref{sec:sec_Hash_Collisions}. With one cover scene and $\Phi$ hidden scenes, the total number of active cells being mapped to the same feature table $T$ is $\kappa_{total} \approx (\Phi+1) \times \kappa$, dramatically increasing the feature interference.

We used the Blender Synthetic dataset for this test, using the $T=2^{19}$ default hash table size. The results in Table \ref{tab:supp_ablation_multi_scene} confirm our hypothesis.

\begin{table}[htbp!]
    \centering
    \caption{Quantitative results for embedding $\Phi$ hidden scenes into a single model, using the default $T=2^{19}$ hash table. As $\Phi$ increases, the increased hash collisions cause a clear degradation in reconstruction fidelity across all scenes.}
    \label{tab:supp_ablation_multi_scene}
    \begin{tabular}{@{}l l c c c@{}}
        \toprule
        \textbf{$\Phi$} & \textbf{Scene} & \textbf{PSNR $\uparrow$} & \textbf{SSIM $\uparrow$} & \textbf{LPIPS $\downarrow$} \\
        \midrule
        \multirow{2}{*}{\textbf{$1$}} & $S$ (Hotdog) & 30.726 & 0.949 & 0.053 \\
         & $\mathcal{B}_1$ (Chair) & 27.508 & 0.919 & 0.090 \\
        \midrule
        \multirow{3}{*}{\textbf{$2$}} & $S$ (Hotdog) & 31.606 & 0.950 & 0.055 \\
         & $\mathcal{B}_1$ (Chair) & 28.141 & 0.919 & 0.090 \\
         & $\mathcal{B}_2$ (Mic) & 28.566 & 0.955 & 0.051 \\
        \midrule
        \multirow{4}{*}{\textbf{$3$}} & $S$ (Hotdog) & 31.037 & 0.946 & 0.063 \\
         & $\mathcal{B}_1$ (Chair) & 27.699 & 0.914 & 0.094 \\
         & $\mathcal{B}_2$ (Mic) & 28.137 & 0.953 & 0.056 \\
         & $\mathcal{B}_3$ (Lego) & 26.496 & 0.900 & 0.083 \\
        \midrule
        \multirow{5}{*}{\textbf{$4$}} & $S$ (Hotdog) & 29.987 & 0.940 & 0.069 \\
         & $\mathcal{B}_1$ (Chair) & 27.331 & 0.907 & 0.105 \\
         & $\mathcal{B}_2$ (Mic) & 27.497 & 0.949 & 0.061 \\
         & $\mathcal{B}_3$ (Lego) & 25.798 & 0.887 & 0.094 \\
         & $\mathcal{B}_4$ (Ship) & 24.663 & 0.800 & 0.204 \\
        \bottomrule
    \end{tabular}
\end{table}


\begin{figure*}[htp]
\scriptsize
\centering
\setlength{\tabcolsep}{1pt}
\begin{tabular}{m{1.7cm}m{1.8cm}m{1.8cm}m{1.8cm}m{1.8cm}m{1.8cm}m{1.8cm}m{1.8cm}m{1.8cm}}

      {}
     &  \multicolumn{2}{c}{$\Phi=1$} 
     &  \multicolumn{2}{c}{$\Phi=2$} 
     &  \multicolumn{2}{c}{$\Phi=3$} 
     &  \multicolumn{2}{c}{$\Phi=4$} \\

     {Hotdog ($S$) } 
     &  \includegraphics[width=1\linewidth]{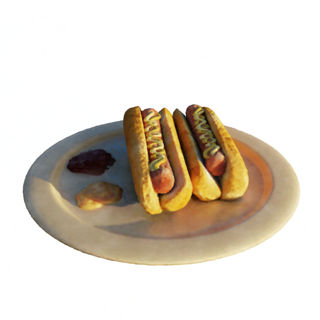}
     &  \includegraphics[width=1\linewidth]{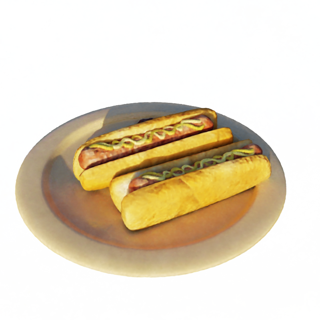}
     &  \includegraphics[width=1\linewidth]{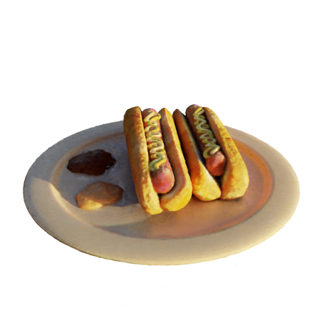}
     &  \includegraphics[width=1\linewidth]{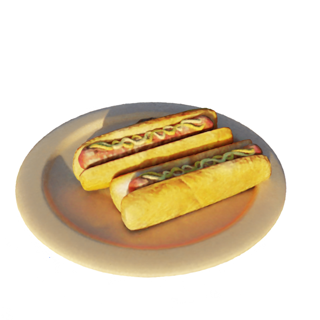}
     &  \includegraphics[width=1\linewidth]{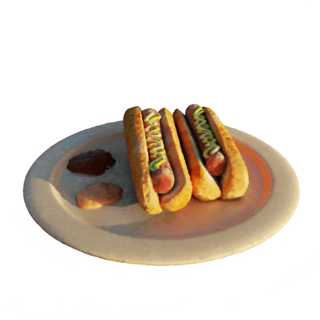}
     &  \includegraphics[width=1\linewidth]{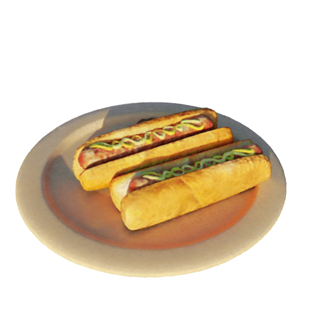}
     &  \includegraphics[width=1\linewidth]{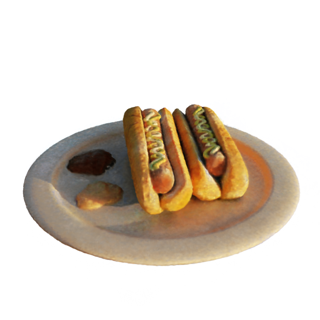}
     &  \includegraphics[width=1\linewidth]{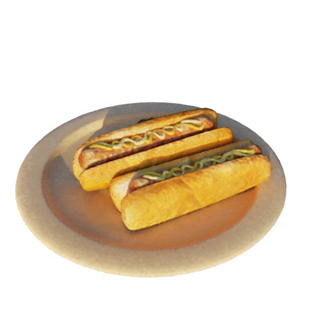} \\

     {Chair ($\mathcal{B}_1$) } 
     &  \includegraphics[width=1\linewidth]{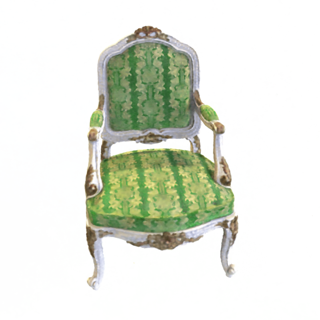}
     &  \includegraphics[width=1\linewidth]{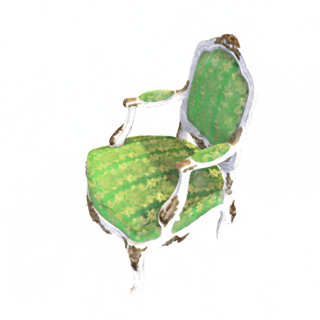}
     &  \includegraphics[width=1\linewidth]{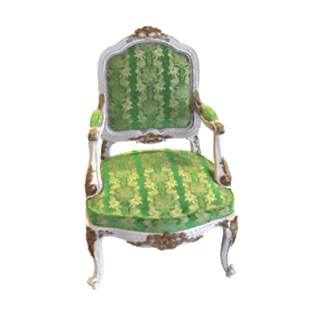}
     &  \includegraphics[width=1\linewidth]{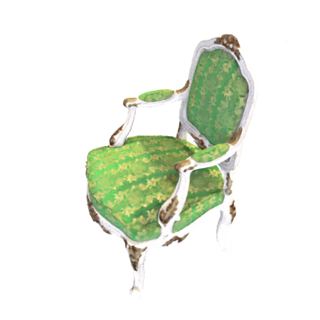}
     &  \includegraphics[width=1\linewidth]{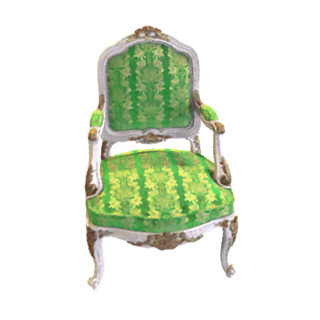}
     &  \includegraphics[width=1\linewidth]{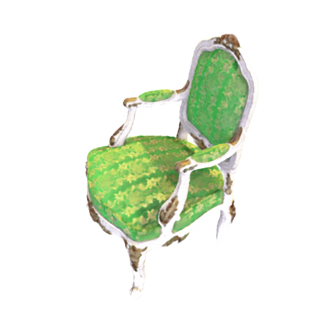}
     &  \includegraphics[width=1\linewidth]{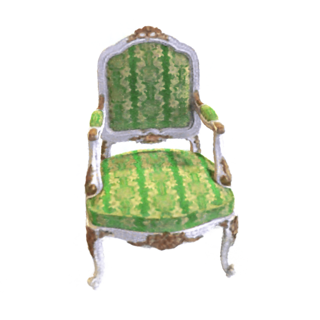}
     &  \includegraphics[width=1\linewidth]{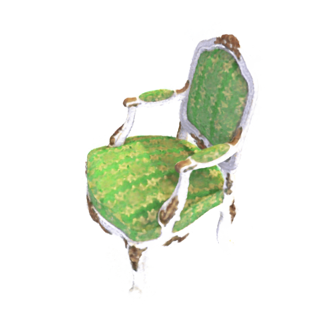} \\ 

     {Mic ($\mathcal{B}_2$) } 
     &  
     &  
     &  \includegraphics[width=1\linewidth]{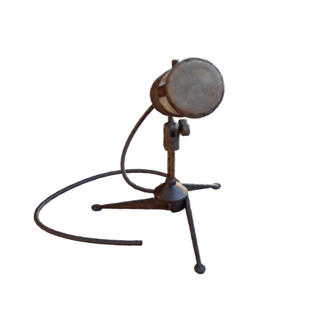}
     &  \includegraphics[width=1\linewidth]{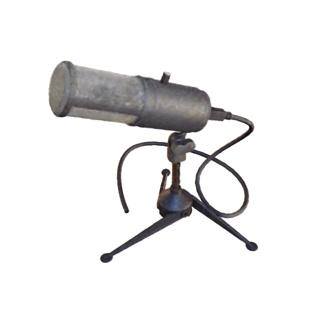}
     &  \includegraphics[width=1\linewidth]{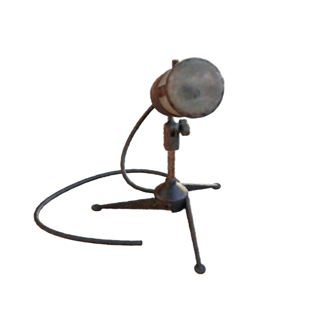}
     &  \includegraphics[width=1\linewidth]{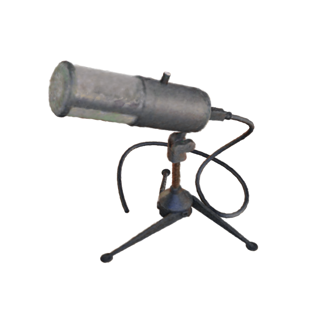}
     &  \includegraphics[width=1\linewidth]{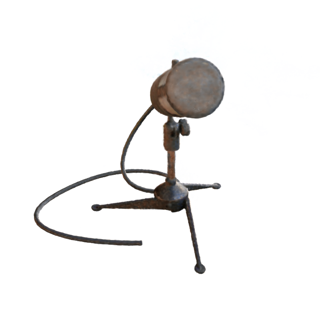}
     &  \includegraphics[width=1\linewidth]{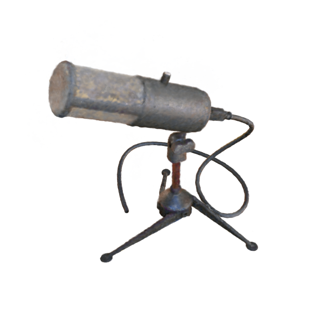} \\ 

     {Lego ($\mathcal{B}_3$) } 
     &  
     &  
     &  
     & 
     &  \includegraphics[width=1\linewidth]{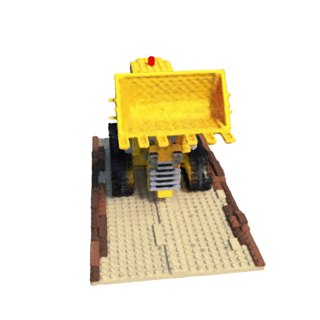}
     &  \includegraphics[width=1\linewidth]{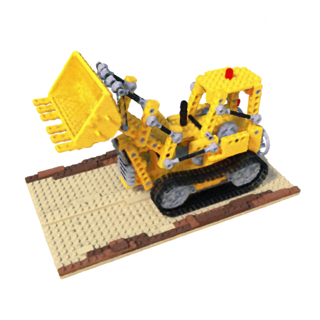}
     &  \includegraphics[width=1\linewidth]{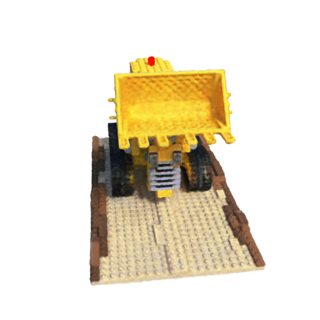}
     &  \includegraphics[width=1\linewidth]{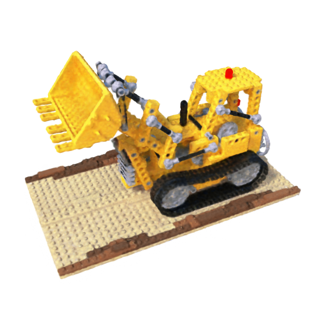} \\ 
     
     {Ship ($\mathcal{B}_4$)} 
     &  
     &  
     &  
     &  
     &  
     &  
     &  \includegraphics[width=1\linewidth]{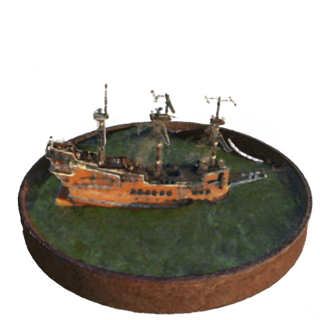}
     &  \includegraphics[width=1\linewidth]{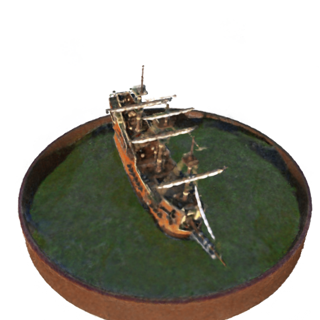} \\ 
    
\end{tabular}
   \caption{Visualization of image results obtained by simultaneously embedding multiple ($\Phi$) hidden scenes ($\mathcal{B}_1$, $\mathcal{B}_2$, $\mathcal{B}_3$ or $\mathcal{B}_4$) and cover scenes ($S$) into a single model using the default $T=2^{19}$ hash table.}
    \label{fig:visual_multi_scenes}
\end{figure*}

\begin{figure*}[htp]
\scriptsize
\centering
\setlength{\tabcolsep}{1pt}
\begin{tabular}{m{1.7cm}m{1.8cm}m{1.8cm}m{1.8cm}m{1.8cm}m{1.8cm}m{1.8cm}m{1.8cm}m{1.8cm}}

      {}
     &  \multicolumn{2}{c}{$\Phi=1$} 
     &  \multicolumn{2}{c}{$\Phi=2$} 
     &  \multicolumn{2}{c}{$\Phi=3$} 
     &  \multicolumn{2}{c}{$\Phi=4$} \\

     {Hotdog ($S$) } 
     &  \includegraphics[width=1\linewidth]{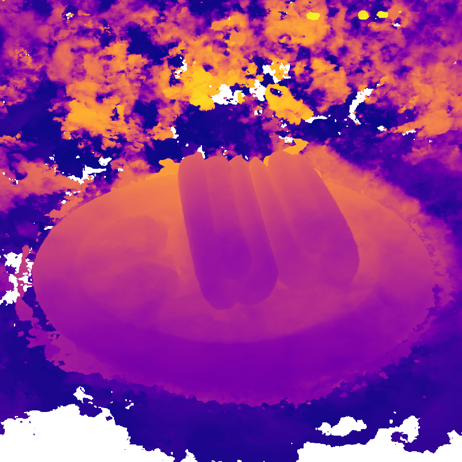}
     &  \includegraphics[width=1\linewidth]{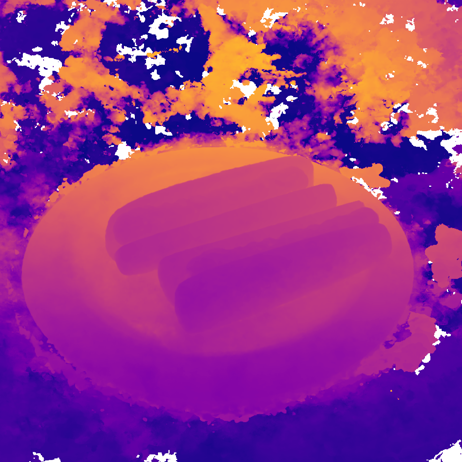}
     &  \includegraphics[width=1\linewidth]{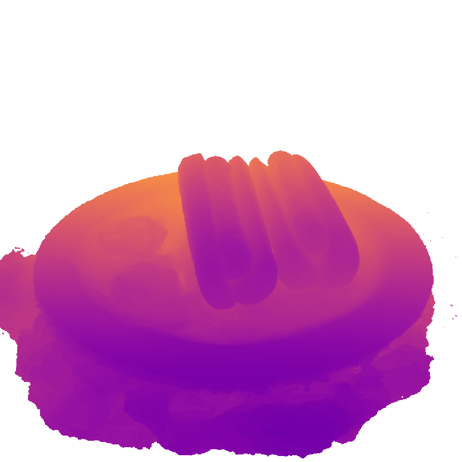}
     &  \includegraphics[width=1\linewidth]{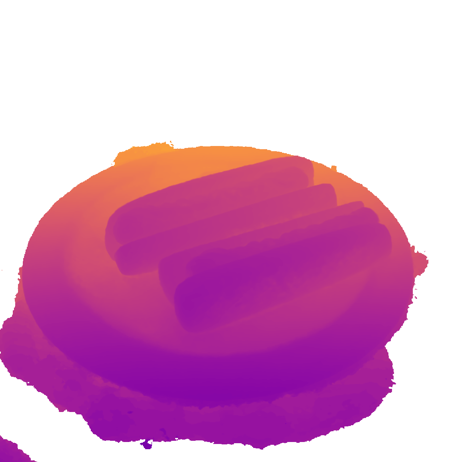}
     &  \includegraphics[width=1\linewidth]{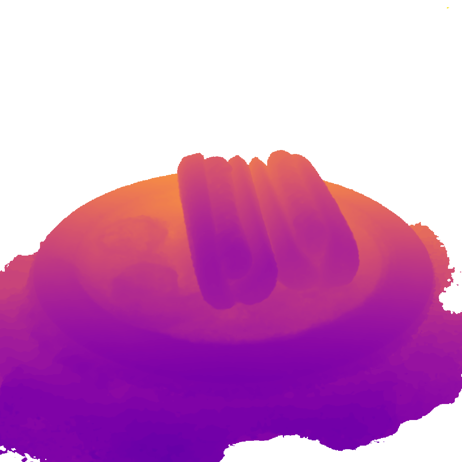}
     &  \includegraphics[width=1\linewidth]{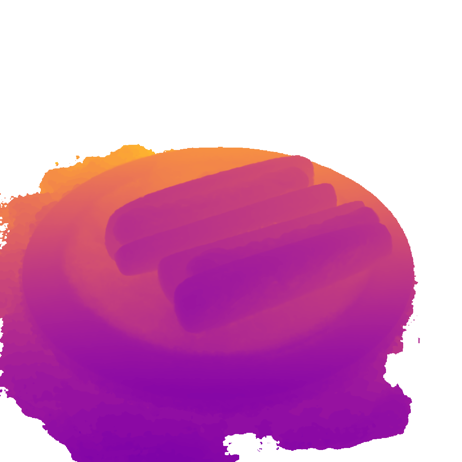}
     &  \includegraphics[width=1\linewidth]{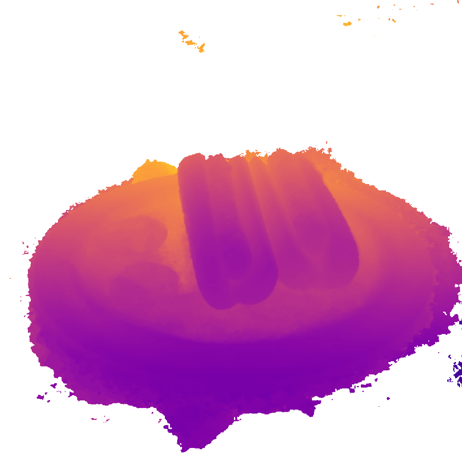}
     &  \includegraphics[width=1\linewidth]{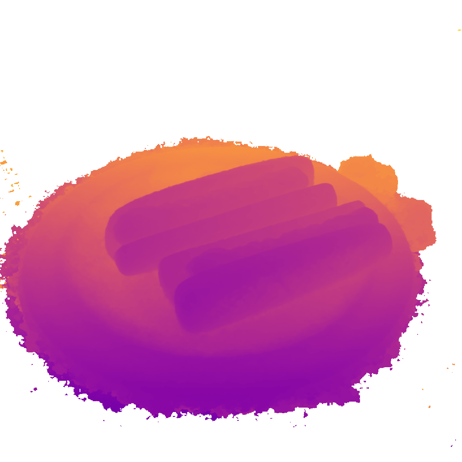} \\

     {Chair ($\mathcal{B}_1$) } 
     &  \includegraphics[width=1\linewidth]{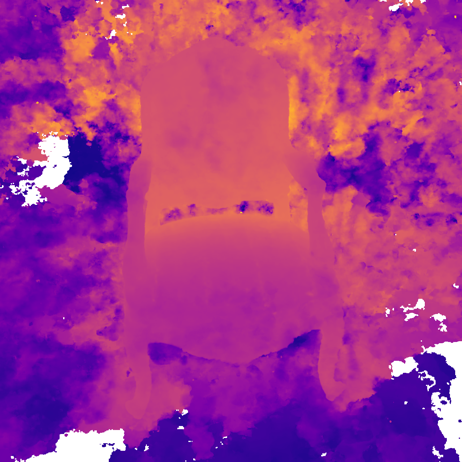}
     &  \includegraphics[width=1\linewidth]{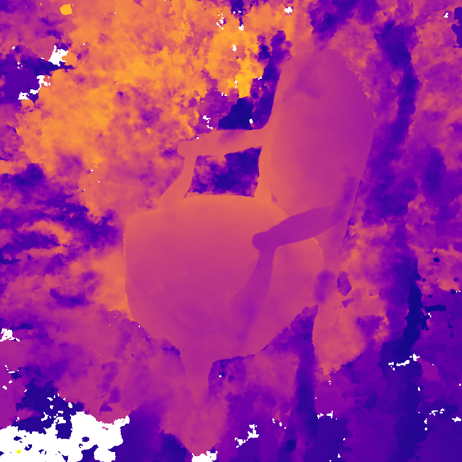}
     &  \includegraphics[width=1\linewidth]{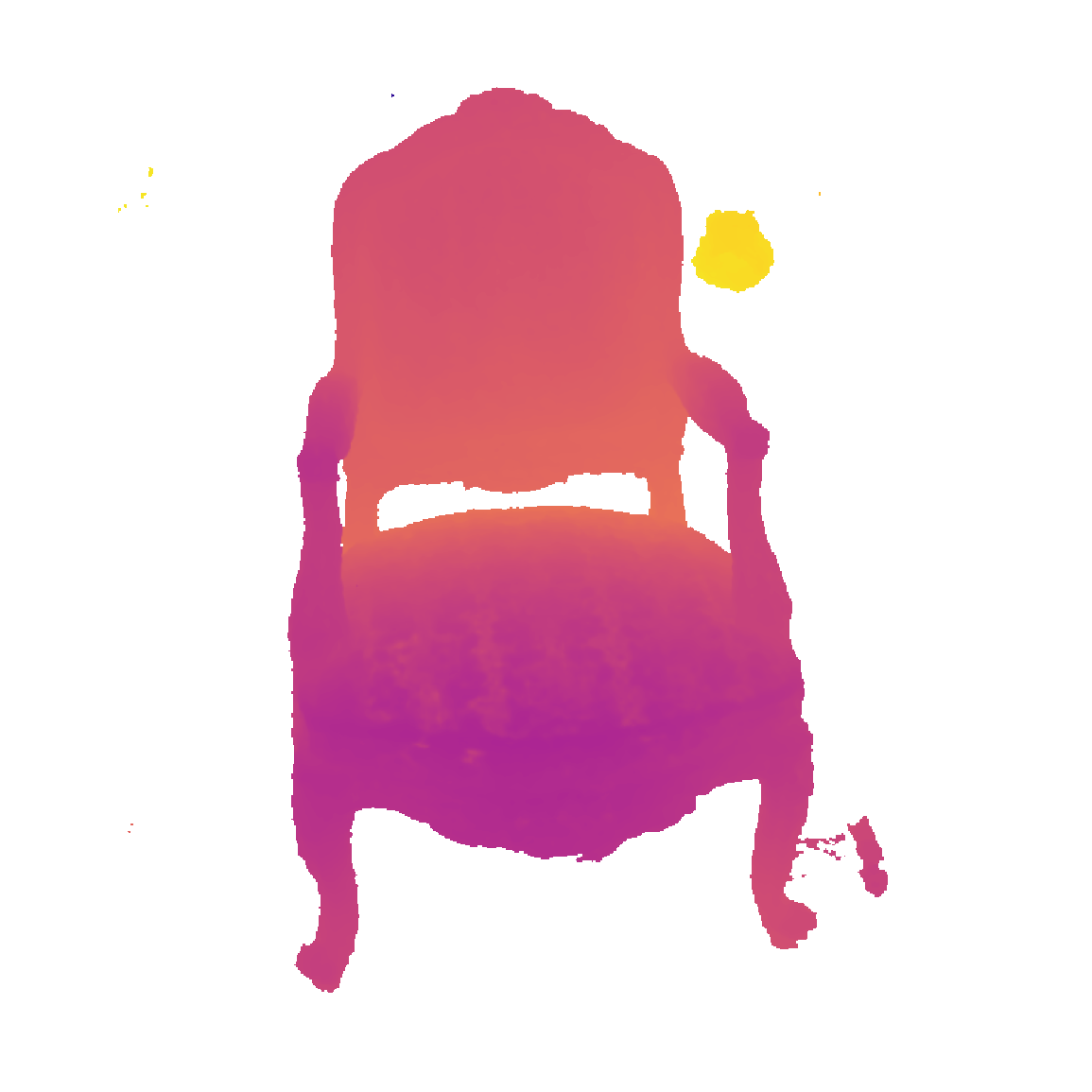}
     &  \includegraphics[width=1\linewidth]{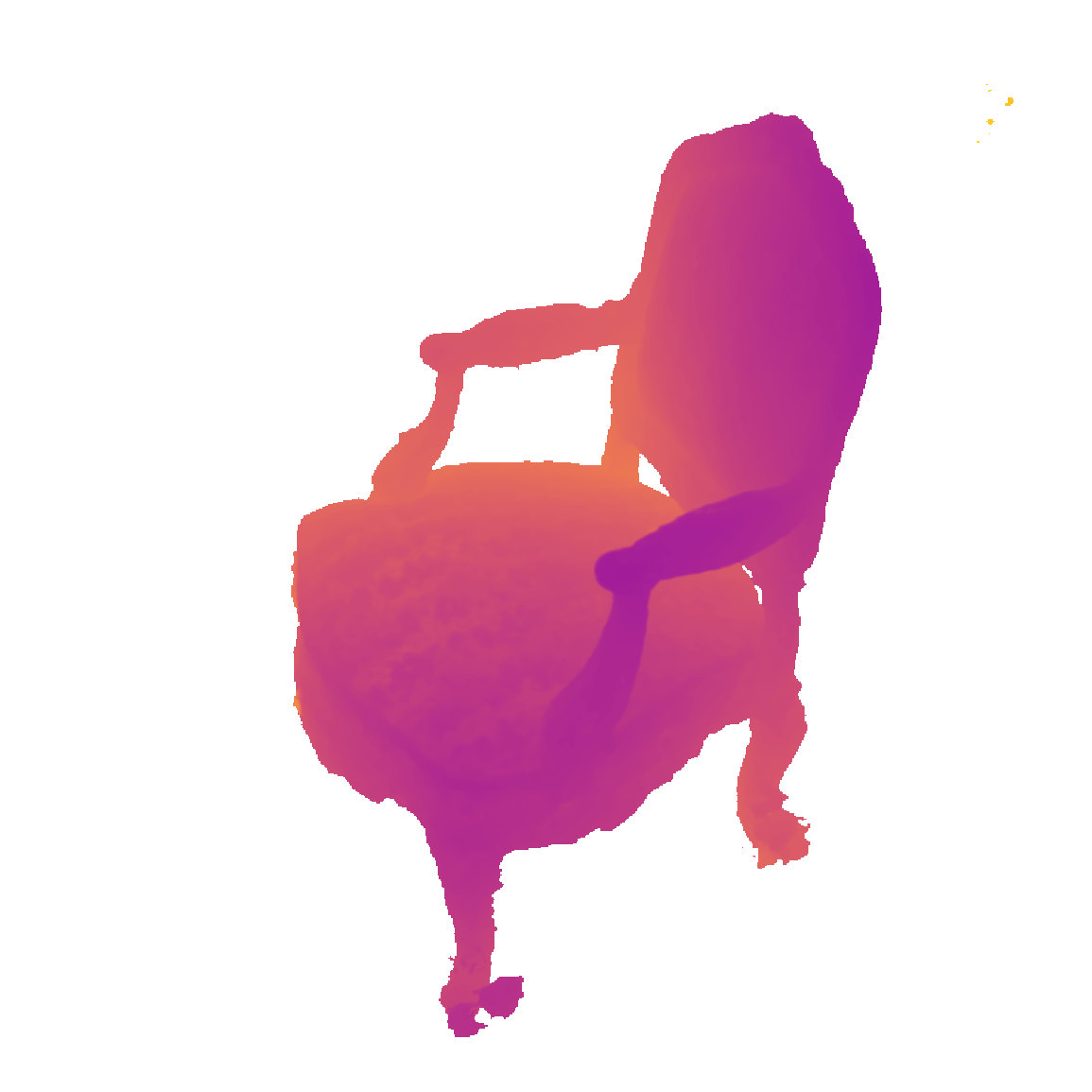}
     &  \includegraphics[width=1\linewidth]{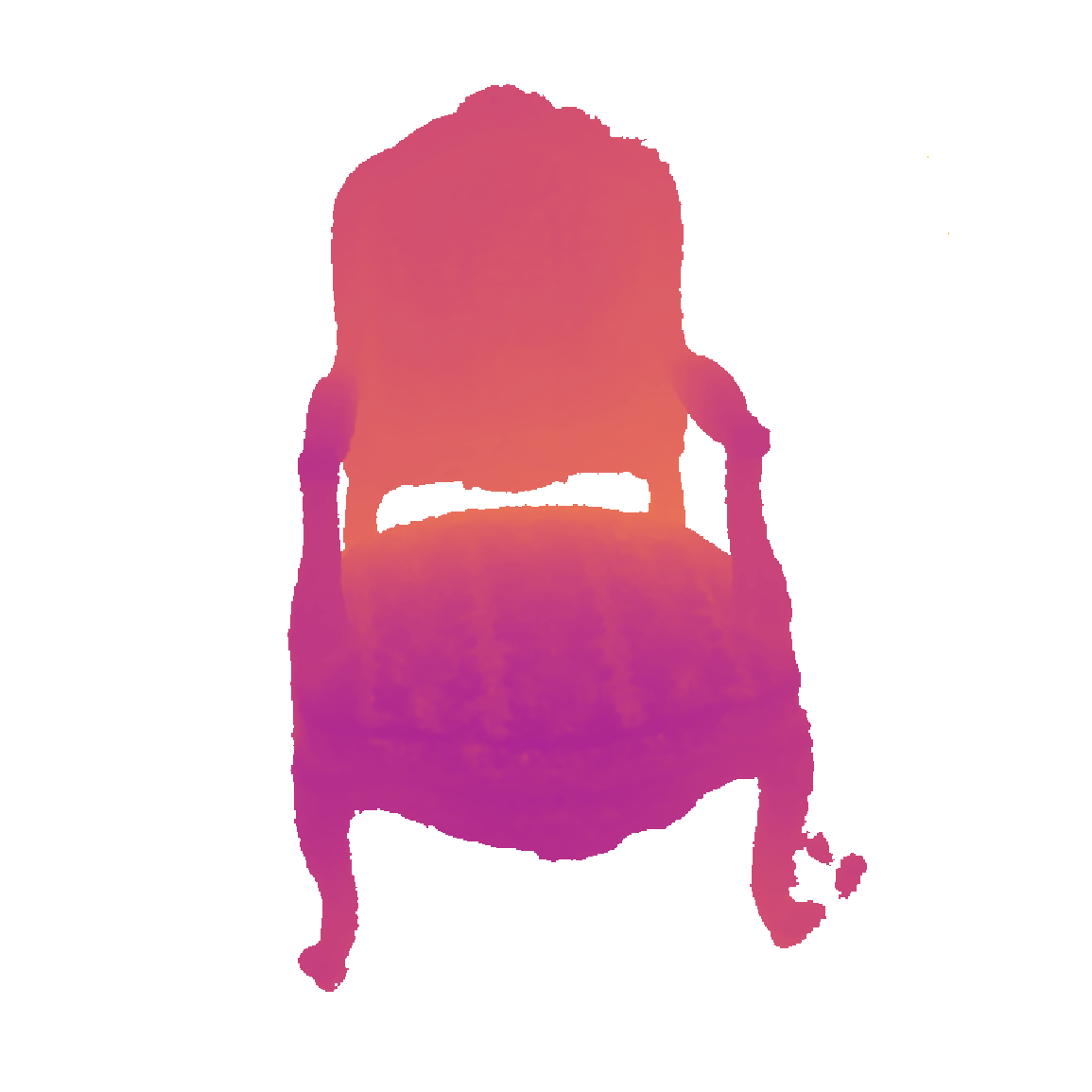}
     &  \includegraphics[width=1\linewidth]{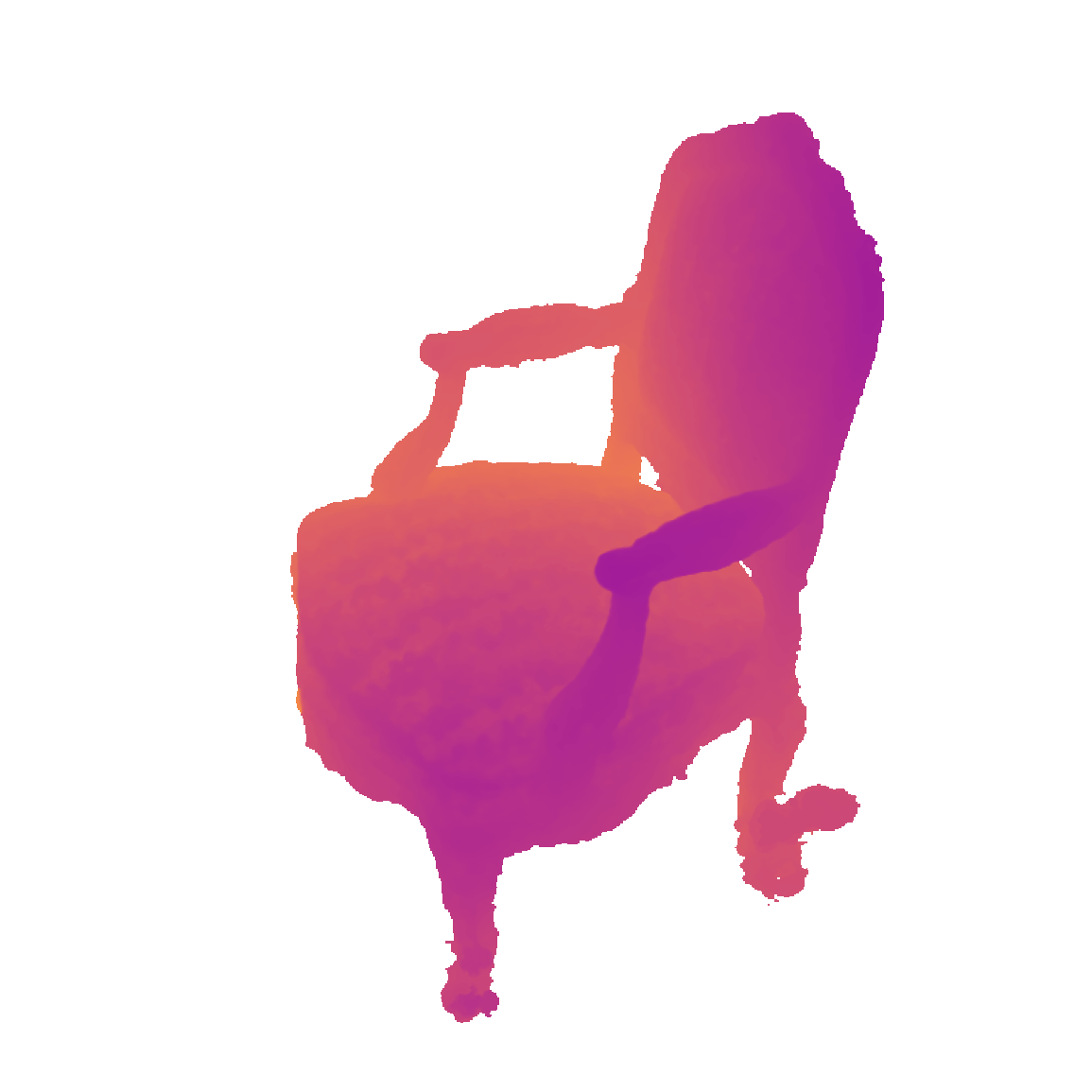}
     &  \includegraphics[width=1\linewidth]{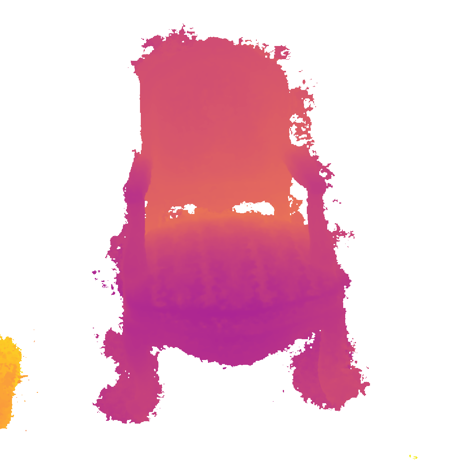}
     &  \includegraphics[width=1\linewidth]{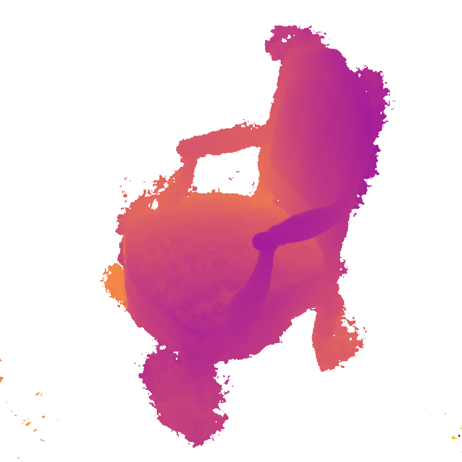} \\ 

     {Mic ($\mathcal{B}_2$) } 
     &  
     &  
     &  \includegraphics[width=1\linewidth]{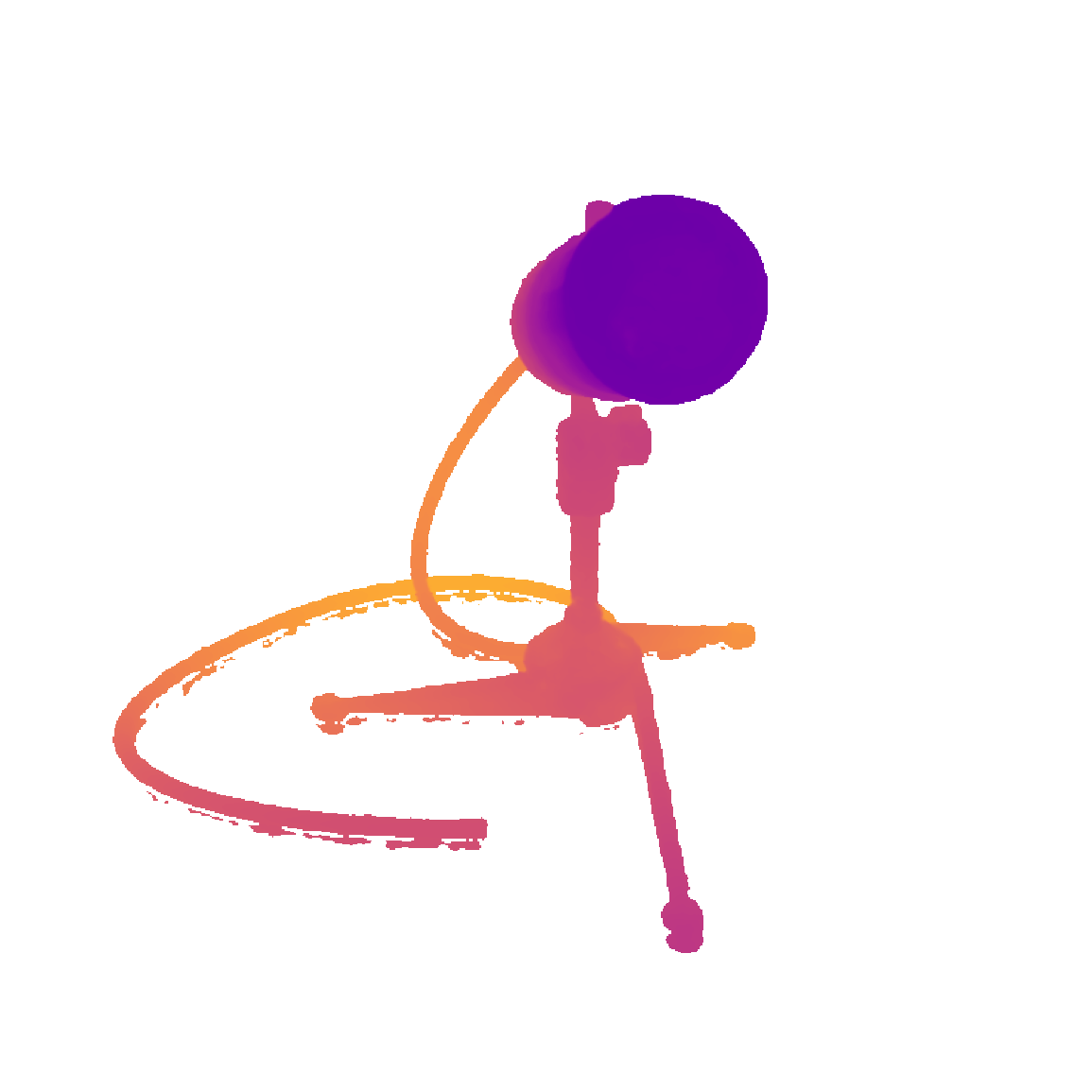}
     &  \includegraphics[width=1\linewidth]{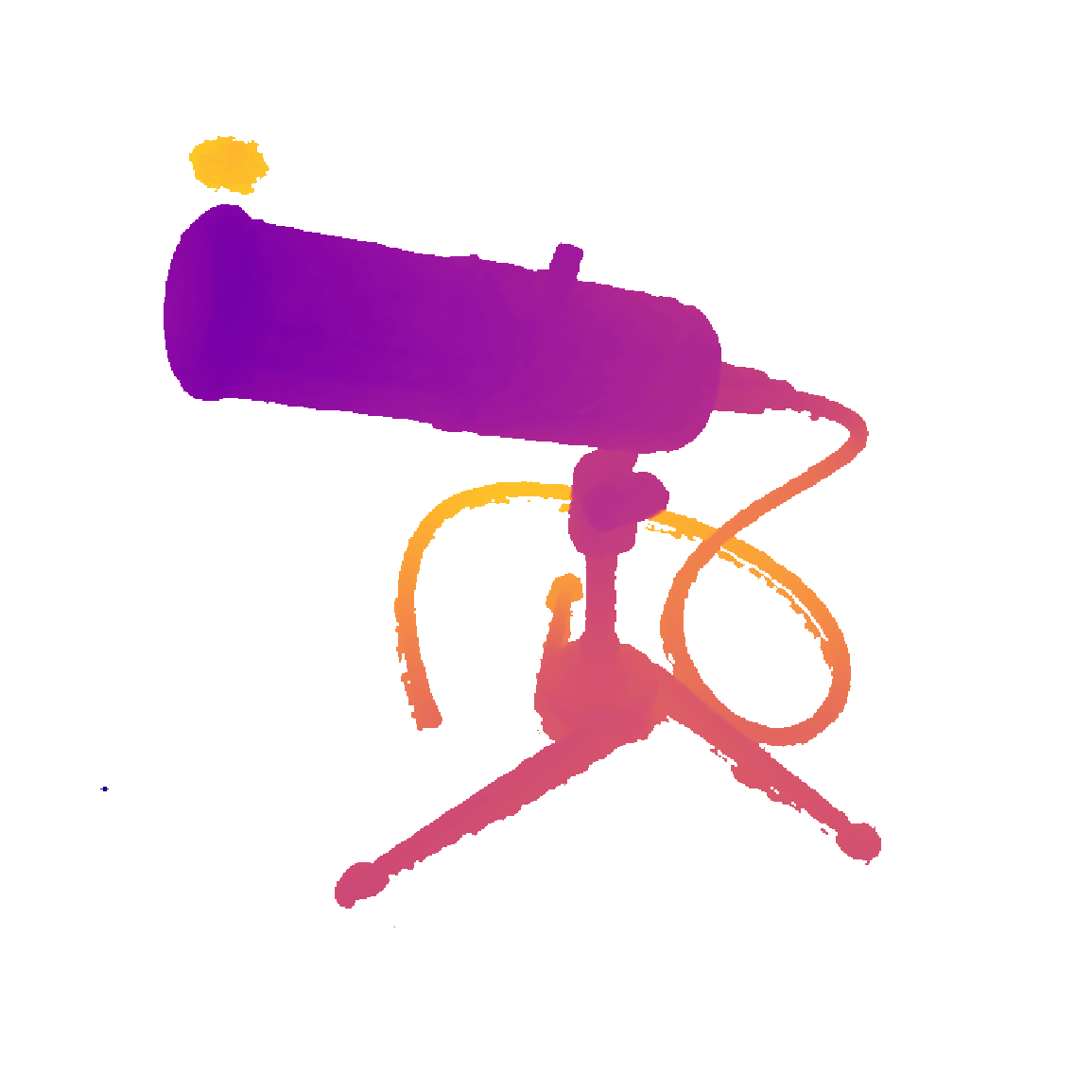}
     &  \includegraphics[width=1\linewidth]{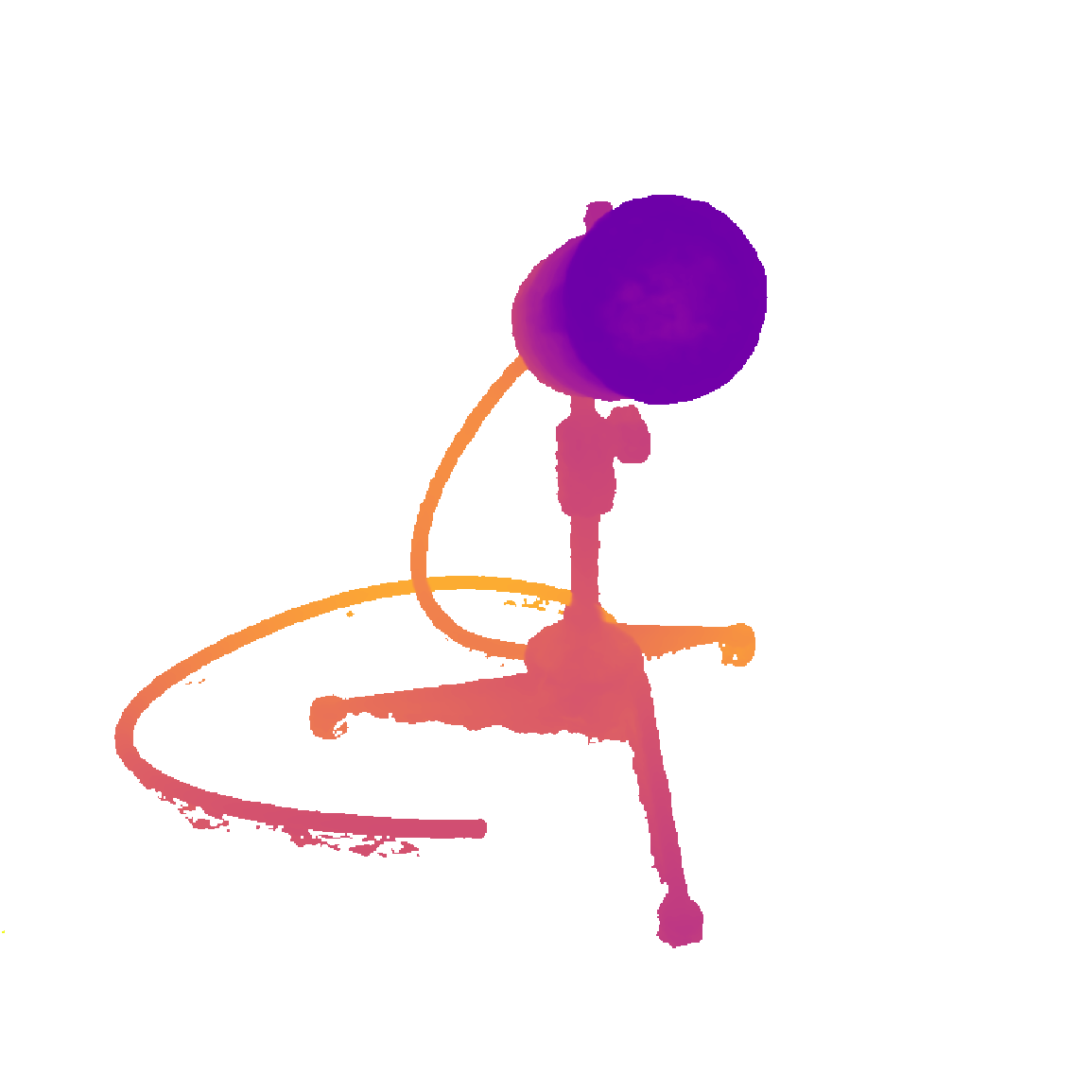}
     &  \includegraphics[width=1\linewidth]{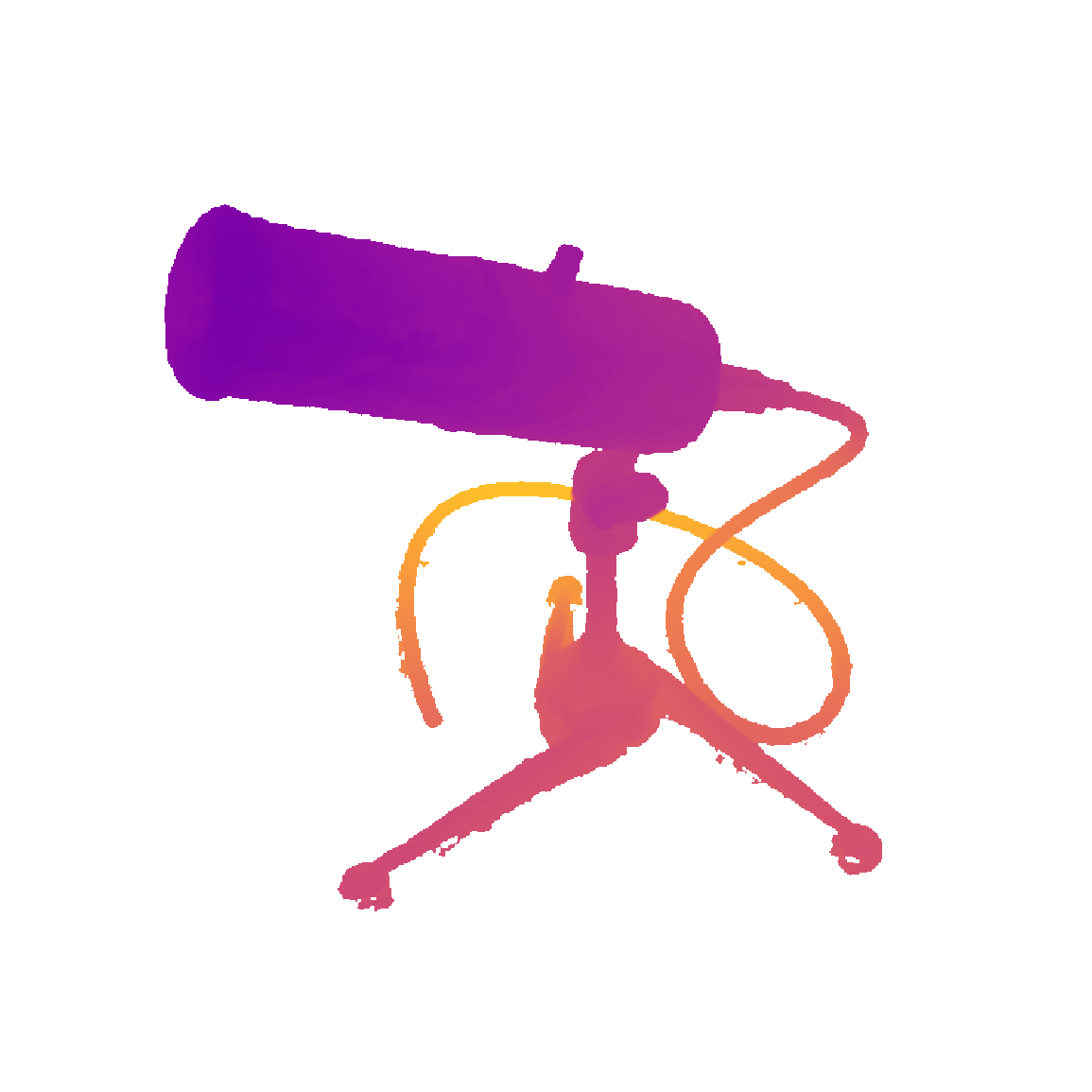}
     &  \includegraphics[width=1\linewidth]{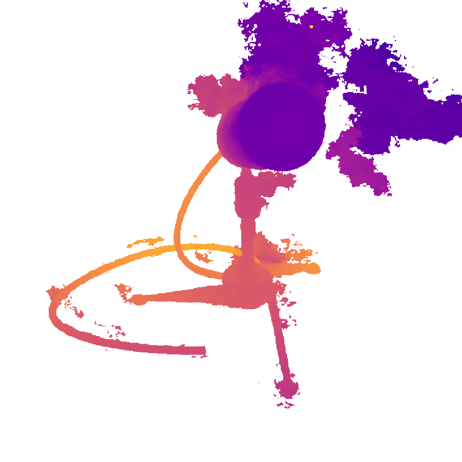}
     &  \includegraphics[width=1\linewidth]{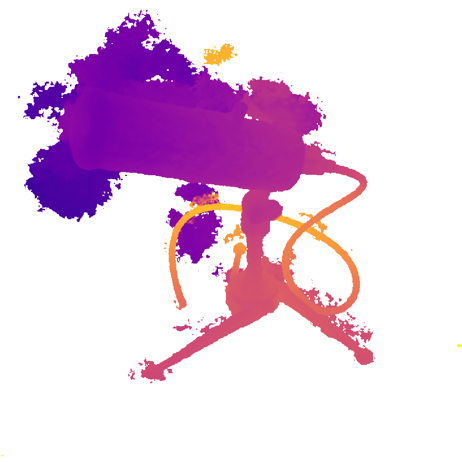} \\ 

     {Lego ($\mathcal{B}_3$) } 
     &  
     &  
     &  
     & 
     &  \includegraphics[width=1\linewidth]{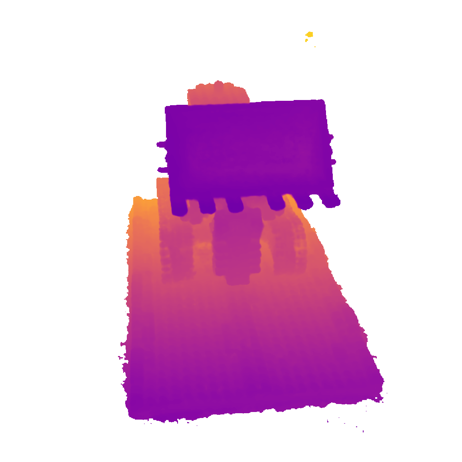}
     &  \includegraphics[width=1\linewidth]{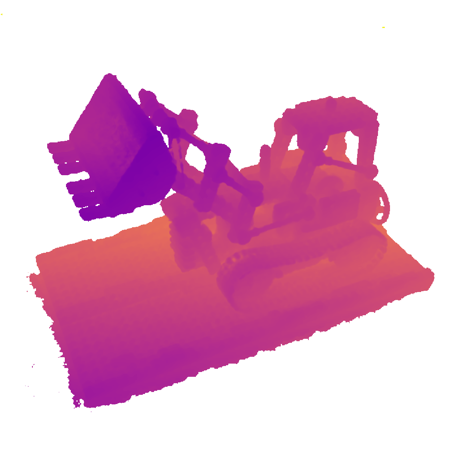}
     &  \includegraphics[width=1\linewidth]{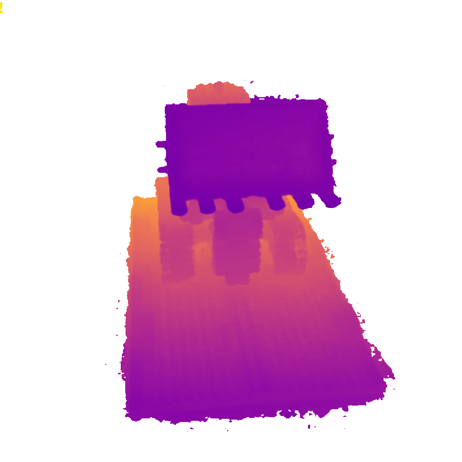}
     &  \includegraphics[width=1\linewidth]{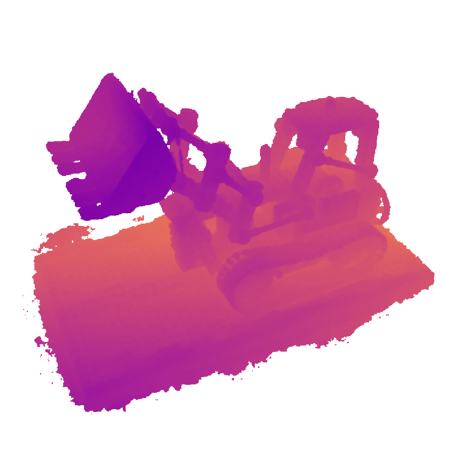} \\ 
     
     {Ship ($\mathcal{B}_4$)} 
     &  
     &  
     &  
     &  
     &  
     &  
     &  \includegraphics[width=1\linewidth]{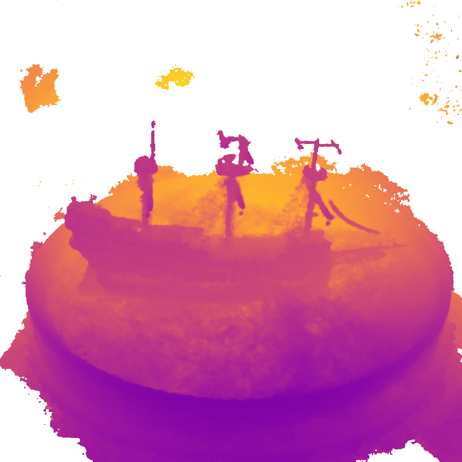}
     &  \includegraphics[width=1\linewidth]{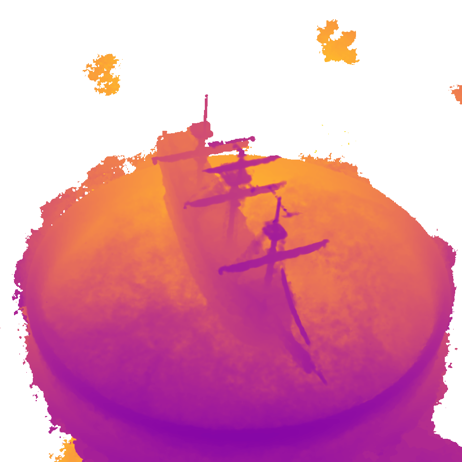} \\ 
    
\end{tabular}
   \caption{Visualization of depth results when embedding multiple ($\Phi$) hidden scenes ($\mathcal{B}_1$, $\mathcal{B}_2$, $\mathcal{B}_3$ or $\mathcal{B}_4$) and cover scenes ($S$) simultaneously into a single model using the default $T=2^{19}$ hash table.}
    \label{fig:visual_multi_scenes_depth}
\end{figure*}

\end{document}